\documentclass[12pt]{article}

\usepackage[a4paper,top=3cm,bottom=2cm,left=2cm,right=2cm,marginparwidth=1.25cm]{geometry}
\usepackage{subfigure}
\usepackage[utf8]{inputenc} 
\usepackage[T1]{fontenc}    
\usepackage{lmodern}
\usepackage{hyperref}       
\usepackage{url}            
\usepackage{booktabs}       
\usepackage{amsfonts}       
\usepackage{nicefrac}       
\usepackage{microtype}      
\newcommand {\myvec}[1] {{\mbox{\boldmath $#1$}}}

\usepackage{amssymb}
\usepackage{amsmath}
\usepackage{mathtools}
\usepackage{amsthm}
\usepackage{color}
\usepackage{xr}
\usepackage{adjustbox}

\DeclareMathOperator{\erf}{erf}
\newcommand{\beginsupplement}{%
        \setcounter{table}{0}
        \renewcommand{\thetable}{S\arabic{table}}%
        \setcounter{figure}{0}
        \renewcommand{\thefigure}{S\arabic{figure}}         \setcounter{section}{0}
        \renewcommand{\thesection}{S\arabic{section}}
}
\makeatletter

\newtheorem{theorem}{Theorem}[section]
\newtheorem{lemma}[theorem]{Lemma}

\makeatletter
\newcommand{\distas}[1]{\mathbin{\overset{#1}{\kern\z@\sim}}}%
\newsavebox{\mybox}\newsavebox{\mysim}
\newcommand{\distras}[1]{%
  \savebox{\mybox}{\hbox{\kern3pt$\scriptstyle#1$\kern3pt}}%
  \savebox{\mysim}{\hbox{$\sim$}}%
  \mathbin{\overset{#1}{\kern\z@\resizebox{\wd\mybox}{\ht\mysim}{$\sim$}}}%
}

\title{ Differentiable Unsupervised Feature Selection based on a Gated Laplacian}

\author{Ofir Lindenbaum $^{1 \ast }$ \and Uri Shaham $^{1 \ast }$ \and Jonathan Svirsky $^{2}$ \and Erez Peterfreund$^{3}$\and Yuval Kluger$^{1 \dagger}$\\
\normalsize{$^{1}$Yale University, USA;}
\normalsize{$^{2}$ Technion, Israel;}
\normalsize{$^{3}$ Hebrew University, Israel}\\
\normalsize{$^\dagger$Corresponding author. E-mail: yuval.kluger@yale.edu}\\\normalsize{Address: 333 Cedar St, New Haven, CT 06510, USA}\\
\normalsize{$^\ast$ These authors contributed equally.}
}

\date{}

\begin{document}

\maketitle

\begin{abstract}
Scientific observations may consist of a large number of variables (features). Identifying a subset of meaningful features is often ignored in unsupervised learning, despite its potential for unraveling clear patterns hidden in the ambient space. In this paper, we present a method for unsupervised feature selection, and we demonstrate its use for the task of clustering. We propose a differentiable loss function which combines the Laplacian score, that favors low frequency features, with a gating mechanism for feature selection. We improve the Laplacian score, by replacing it with a gated variant computed on a subset of features. This subset is obtained using a continuous approximation of Bernoulli variables whose parameters are trained to gate the full feature space. We mathematically motivate the proposed approach, and demonstrate that in the high noise regime, it is crucial to compute the Laplacian on the gated inputs, rather than on the full feature set. Experimental demonstration of the efficacy of the proposed approach and its advantage over current baselines is provided using several real-world examples.
\end{abstract}

\section{Introduction}

Recently there has been a growing interest in the machine learning community towards unsupervised and self-supervised learning. This was motivated by impressive empirical results demonstrating the benefits of analyzing large amounts of unlabeled data (for example in natural language processing). 
In many scientific domains, such as biology and physics, the growth of computational and storage resources, as well as technological advances for measuring numerous features simultaneously, makes the analysis of large, high dimensional datasets an important research need. 
In such datasets, discarding irrelevant (i.e. noisy and information-poor) features may reveal clear underlying natural structures that are otherwise hidden in the high dimensional space.
We refer to these uninformative features as ``nuisance features''. 
While nuisance features are mildly harmful in the supervised regime, in the unsupervised regime discarding such features is crucial and may determine the success of downstream analysis tasks (e.g., clustering or manifold learning). 
Some of the pitfalls caused by nuisance features could be mitigated using an appropriate unsupervised feature selection method.

The problem of feature selection has been studied extensively in machine learning and statistics. Most of the research is focused on supervised feature selection, where identifying a subset of informative features has benefits such as reduction in memory and computations, improved generalization and interpretability.   
Filter methods, such as \cite{MI1,MI2,MI3,HSIC1,HSIC2,HSIC3} attempt to remove irrelevant features prior to learning a model. Wrapper methods \cite{wrapper1,wrapper2,wrapper3,wrapper4,KernelW} use the outcome of a model to determine the relevance of each feature. Embedded methods, such as \cite{Lasso,yamada2018feature,DFS,sparseNN-group-lasso,one-layer} aim to learn the model while simultaneously select the subset of relevant features. 

 Unsupervised feature selection methods mostly focus on two main tasks: clustering and dimensionality reduction or manifold learning. Among studies which tackle the former task,~\cite{AE1,AE2,CAE} use autoencoders to identify features that are sufficient for reconstructing the data. Other clustering-dedicated unsupervised feature selection methods asses the relevance of each feature based on different statistical or geometric measures. Entropy, divergence and mutual information are used in \cite{UMI1,UMI2,UMI3,UMI4,UMI5} to identify features which are informative for clustering the data. A popular tool for evaluating features is the graph Laplacian \cite{ng2002spectral,belkin2002laplacian}. The Laplacian Score (LS) \cite{he2006laplacian}, evaluates the importance of each feature by its ability to preserve local structure. The features which most preserve the manifold structure (captured by the Laplacian) are retained. Several studies, such as \cite{zhao2007spectral,padungweang2009univariate,zhao2011spectral}, extend the LS based on different spectral properties of the Laplacian.
 
 While these methods are widely used in the feature selection community, they rely on the success of the Laplacian in capturing the ``true'' structure of the data. 
 We argue that when the Laplacian is computed based on all features, it often fails to identify the informative ones. This may happen in the presence of a large number of nuisance features, when the variability of the nuisance features masks the variability associated with the informative features. Scenarios like this are prevalent in areas such as bioinformatics, where a large number of biomarkers are measured to characterize developmental and chronological biology processes such as cell differentiation or cell cycle.
 These processes may depend merely on a few biomarkers. In these situations, it is desirable to have an unsupervised method that can filter nuisance features prior to the computation of the Laplacian.
  
 In this study, we propose a differentiable objective for unsupervised feature selection. 
 Our proposed method utilizes trainable stochastic input gates, trained to select features with high correlation with the leading eigenvectors of a graph Laplacian that is computed based on these features. This gating mechanism allows us to re-evaluate the Laplacian for different subsets of features and thus unmask informative structures buried by the nuisance features. We demonstrate, that the proposed approach outperforms several unsupervised feature selection baselines.

\section{Preliminaries}
\label{sec:background}
Consider a data matrix $\myvec{X} \in \mathbb{R}^{n \times d}$, with $d$ dimensional observations $\myvec{x}_1,...,\myvec{x}_n$. We refer to the columns of $\myvec{X}$ as features $\myvec{f}_1,...,\myvec{f}_d$, where  $\myvec{f}_i\in \mathbb{R}^n$ and assume that features are centered and normalized such that $\myvec{1}^T\myvec{f}_i = 0 $ and $\|\myvec{f}_i \|^2_2=1$. 
We assume that the data has an inherent structure, determined by a subset of the features ${\cal{S}^*}$ and that other features are nuisance variables. Our goal is to identify the subset of relevant features ${\cal{S}^*}$ and discard the remaining ones. 

\subsection{Graph Laplacians}
Given $n$ data points, a kernel matrix is a $n \times n$ matrix $\myvec{K}$ so that $\myvec{K}_{i,j}$ represents the similarity between $\myvec{x}_i$ and $\myvec{x}_j$. A popular choice for such matrix based on the Gaussian kernel
    $$\myvec{K}_{i,j} =\exp\left(-\frac{\| \myvec{x}_i-\myvec{x}_j \|^2} {2\sigma_b^2}\right), $$
    where $\sigma_b$ is a user-defined bandwidth (chosen, for example, based on the minimum values of the 1-nearest neighbors of all points).
The unnormalized graph Laplacian matrix is defined as $\myvec{L}_\text{un}=\myvec{D}-\myvec{K}$, where $\myvec{D}$ is a diagonal matrix $\myvec{D}$, whose elements $\myvec{D}_{i,i}=\sum^n_{j=1} \myvec{K}_{i,j}$ correspond to the degrees of the points $i=1,...,n$.
The random walk graph Laplacian is defined as $\myvec{L}_\text{rw}= \myvec{D}^{-1}\myvec{K}$, and expresses the transition probabilities of a random walk to move between data points.
Graph Laplacian matrices are extremely useful in many unsupervised machine learning tasks. In particular, it is known that the eigenvectors corresponding to the small eigenvalues of the unnormalized Laplacian (or the large eigenvalues of the random walk Laplacian) are useful for embedding the data in low dimension (see, for example,~\cite{von2007tutorial}).

\subsection{Laplacian Score}
\label{sec:LS}
Following the success of the graph Laplacian \cite{belkin2002laplacian} and \cite{ng2002spectral}, the authors in \cite{he2006laplacian} have presented an unsupervised measure for feature selection, termed Laplacian Score (LS). 
The LS evaluates each feature based on its correlation with the leading eigenvectors of the graph Laplacian.

At the core of the LS method, the score of feature $\myvec{f}$ is determined by the quadratic form $\myvec{f}^T \myvec{L} \myvec{f}$, where $ \myvec{L} =  \myvec{L}_\text{un}$ is the unnormalized graph Laplacian.
Since 
$$\myvec{f}^T \myvec{L} \myvec{f} = \sum_{i=1}^n\lambda_i\langle \myvec{u}_i, \myvec{f} \rangle^2,$$
where $ \myvec{L} =  \sum_{i=1}^n\lambda_i \myvec{u}_i\myvec{u}_i^T$ is the eigen-decomposition of $\myvec{L}$,  the score is smaller when $\myvec{f}$ has a larger component in the subspace of the smallest eigenvectors of $\myvec{L}$. Such features can be thought of as ``informative'', as they respect the graph structure. Eigenvalues of the Laplacian can be interpreted as frequencies, and eigenvectors corresponding to larger eigenvalues of $\myvec{L}_\text{un}$ (or smaller eigenvalues of $\myvec{L}_\text{rw}$) oscillate faster. Based on the assumption that the interesting underlying structure of the data (e.g. clusters) depends on the  slowly varying features in the data, \cite{he2006laplacian} proposed to select the features with the smallest scores.

\begin{figure*}[htb!] 
\centering
\includegraphics[width=0.3\textwidth]{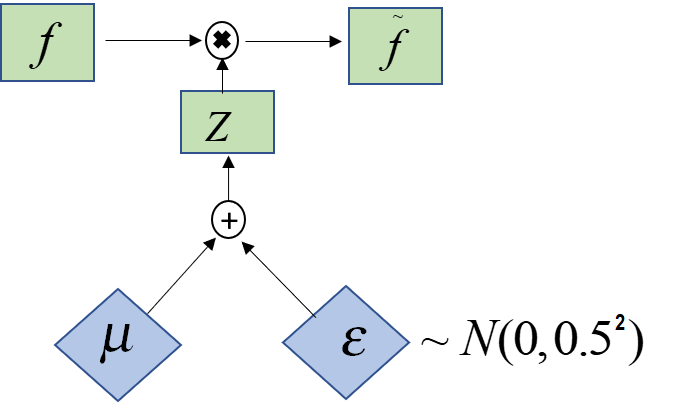}
\includegraphics[width=0.3\textwidth]{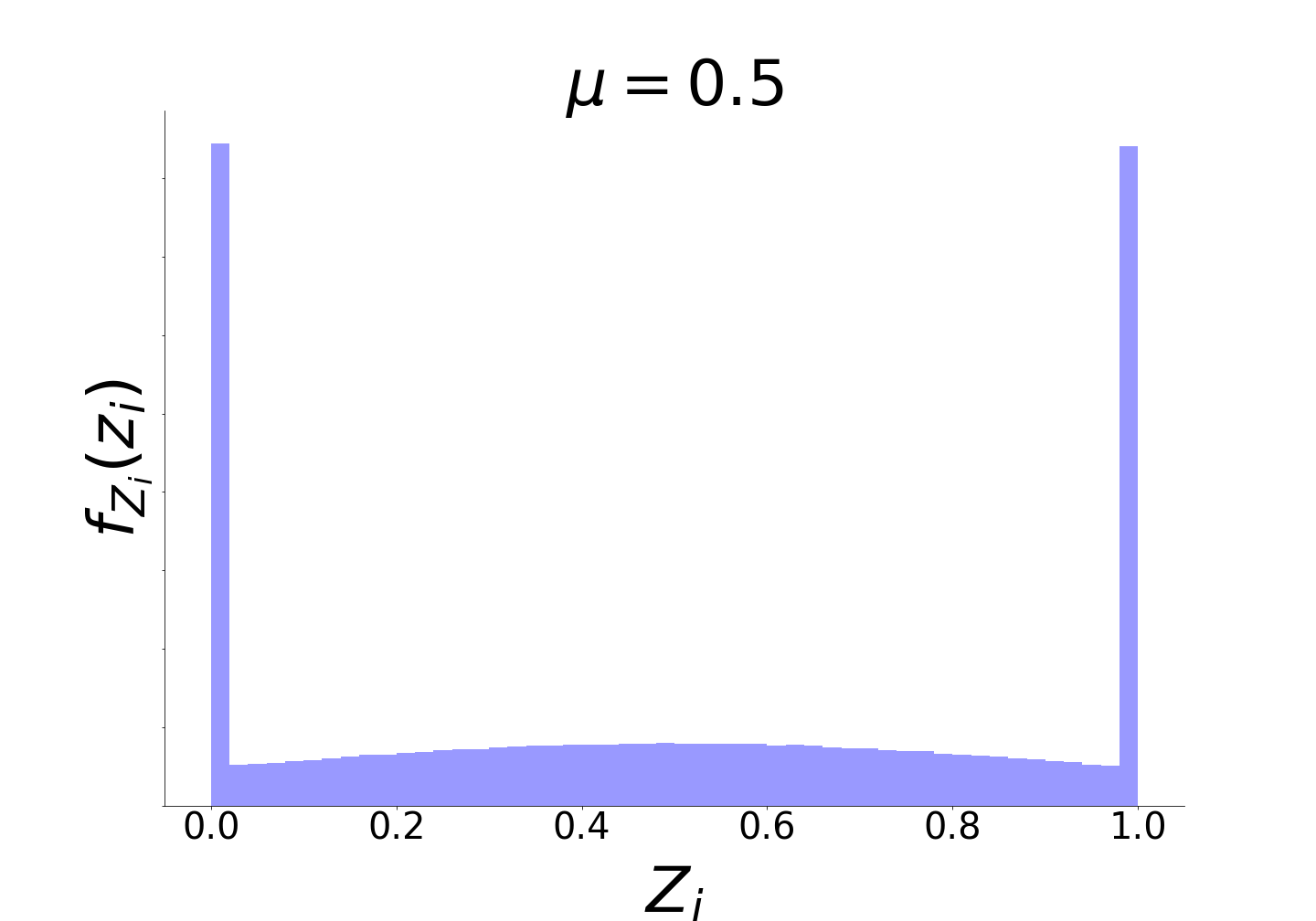} 
\includegraphics[width=0.3\textwidth]{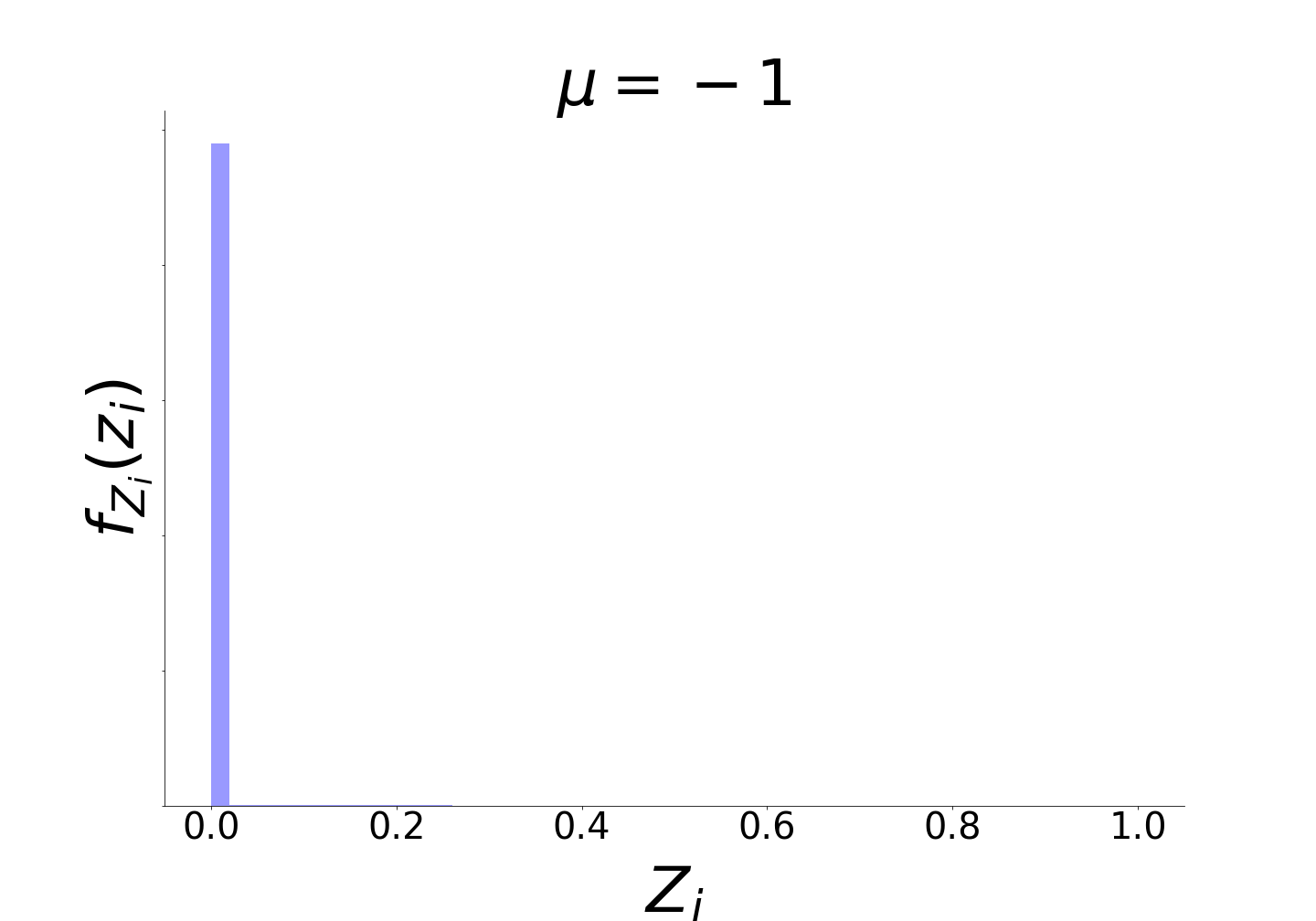} 
\caption{Left: The stochastic gate $Z$ is defined via the repamaterization trick \cite{reparameterization1,reparameterization2}. Standard Gaussian noise is injected and shifted by a trainable parameter $\mu_i$, the result is thresholded to $[0,1]$ based on~\eqref{eq:stg}. Two examples of the density of the stochastic gate $Z_i$. Middle: at initialization $\mu_i=0.5$ and the gate approximates a 'fair' Bernoulli variable. Right: the distribution at $\mu_i=-1$ approximates a 'closed' gate.}
\label{fig:density}
\end{figure*}

\subsection{Stochastic Gates}
\label{sec:stg}
Due to the enormous success of gradient decent-based methods, most notably in deep learning, it is appealing to try to incorporate discrete random variables into a differentiable loss functions designed to retain the slow varying features in the data. However, the gradient estimates of discrete random variables tend to suffer from high variance \cite{he2004locality}. Therefore, several authors have proposed continuous approximations of discrete random variables \cite{maddison2016concrete,Gumbel1}. Such relaxations have been used for many applications, such as model compression \cite{louizos2017learning}, discrete softmax activations \cite{jang2016categorical} and feature selection \cite{yamada2018feature}. Here, we use a Gaussian-based relaxation of Bernoulli variables, termed Stochastic Gates (STG) \cite{yamada2018feature} which is differentiated based on the repamaterization trick \cite{reparameterization1,reparameterization2}.

We denote the STG random vector by $\myvec{Z} \in [0,1]^d$, parametrized by $\mu\in\mathbb{R}^d$, where each entry is defined as 
\begin{equation}\label{eq:stg}
  {Z}_i = \max(0,
 \min(1, \mu_i + \epsilon_i )) , 
\end{equation} where $\mu_i$ is a learnable parameter, $\epsilon_i$ is drawn from  $\mathcal{N}(0 ,\sigma^2)$ and $\sigma$ is fixed throughout training. This approximation
 can be viewed as a clipped, mean-shifted, Gaussian random variable. 
 In Fig.~\ref{fig:density} we illustrate the gating mechanism, and show examples of the densities of ${Z}_i$ for different values of $\mu_i$.

Multiplication of each feature by its corresponding gate enables us to derive a fully differentiable feature selection method. At initialization $\mu_i=0.5,i=1,...,d$, so that all gates approximate a ''fair'' Bernoulli variable.
 The parameters $\mu_i$ can be learned via gradient decent by incorporating the gates in a diffrentiable loss term. To encourage feature selection in the supervised setting, \cite{yamada2018feature} proposed the following differentiable regularization term \begin{equation}
 r(\myvec{Z}) = \sum_{i=1}^d \mathbb{P}({Z}_i \ge 0) = \sum_{i=1}^d\left(\frac{1}{2} - \frac{1}{2} \erf\left(-\frac{\mu_i}{\sqrt{2}\sigma}\right) \right), \label{eq:reg_term}
 \end{equation}
 where $\erf()$ is the Gauss error function. The term \eqref{eq:reg_term} penalizes open gates, so that gates corresponding to features that are not useful for prediction are encouraged to transition into a closed state (which is the case for small $\mu_i$).



\begin{figure*}[htb!]
\centering
\includegraphics[width=0.4\textwidth] {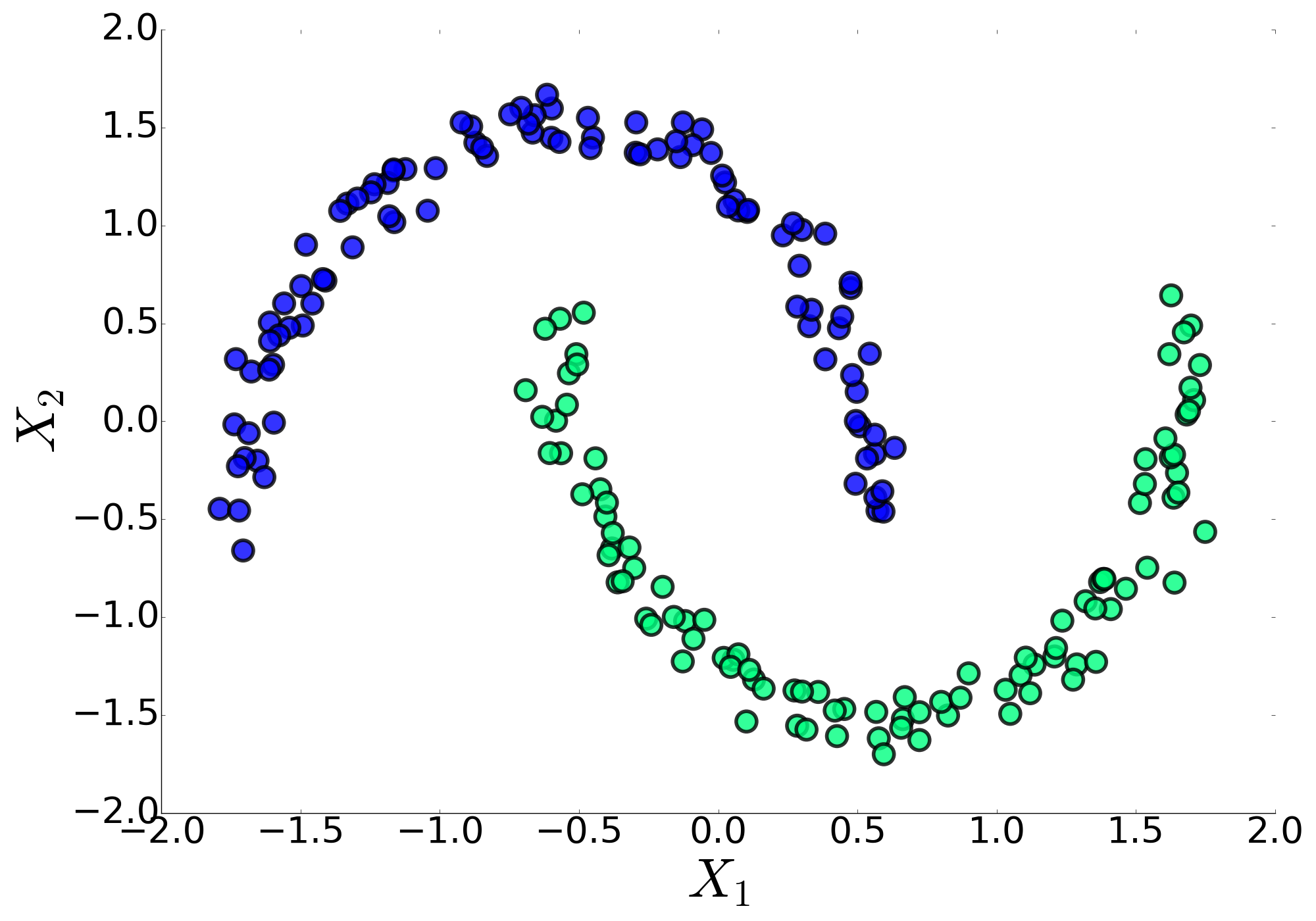}
\includegraphics[width=0.4\textwidth] {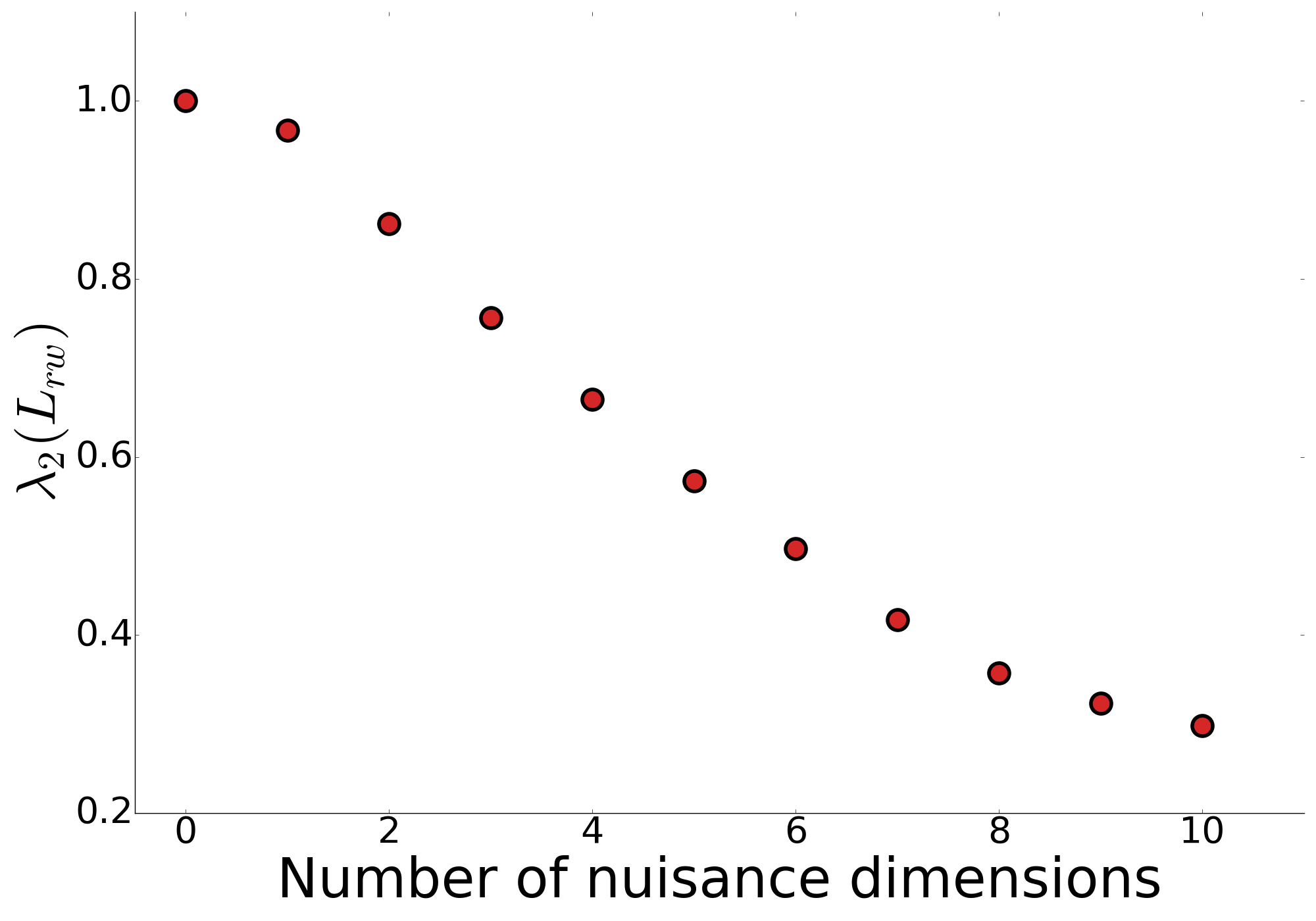}
\includegraphics[width=0.4\textwidth] {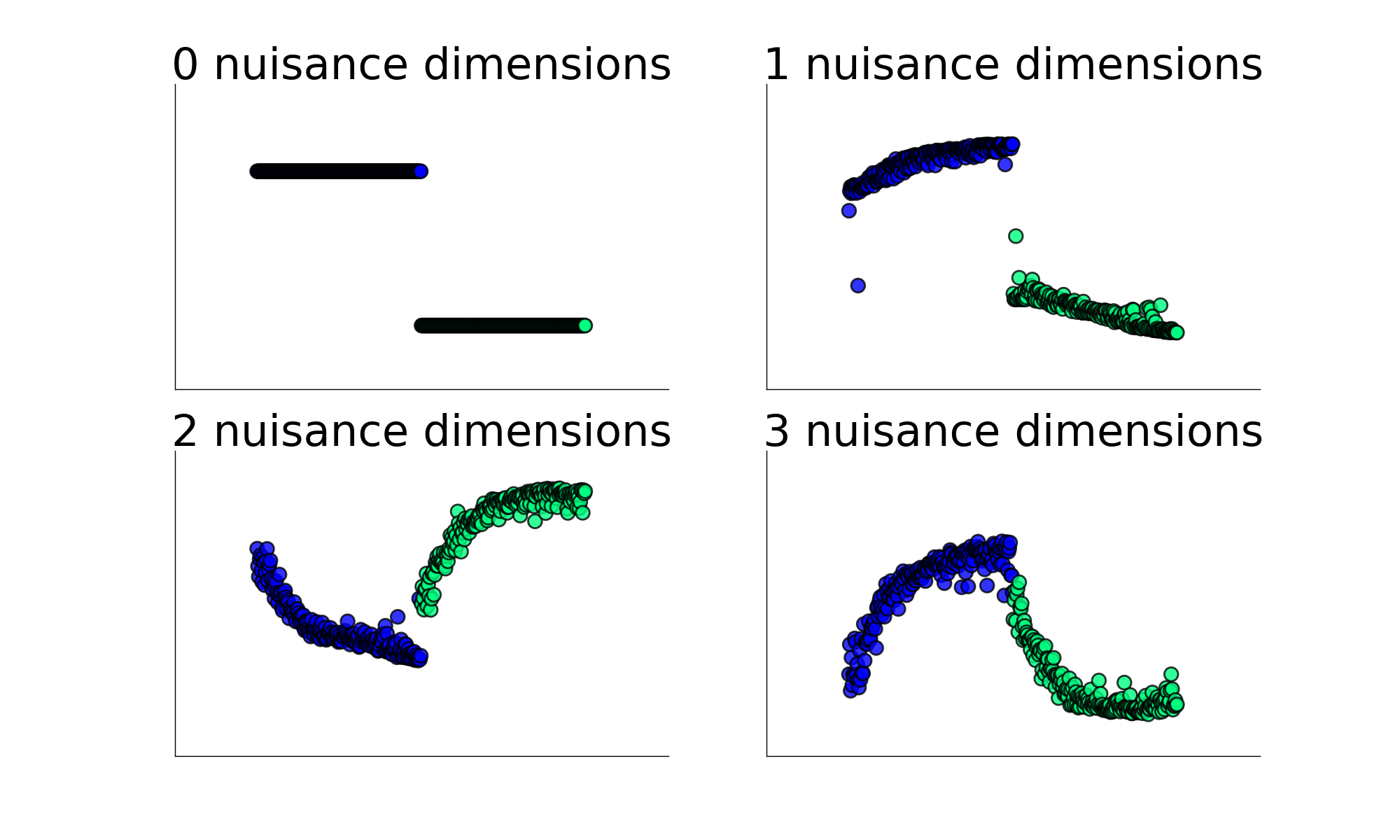}
\includegraphics[width=0.4\textwidth] {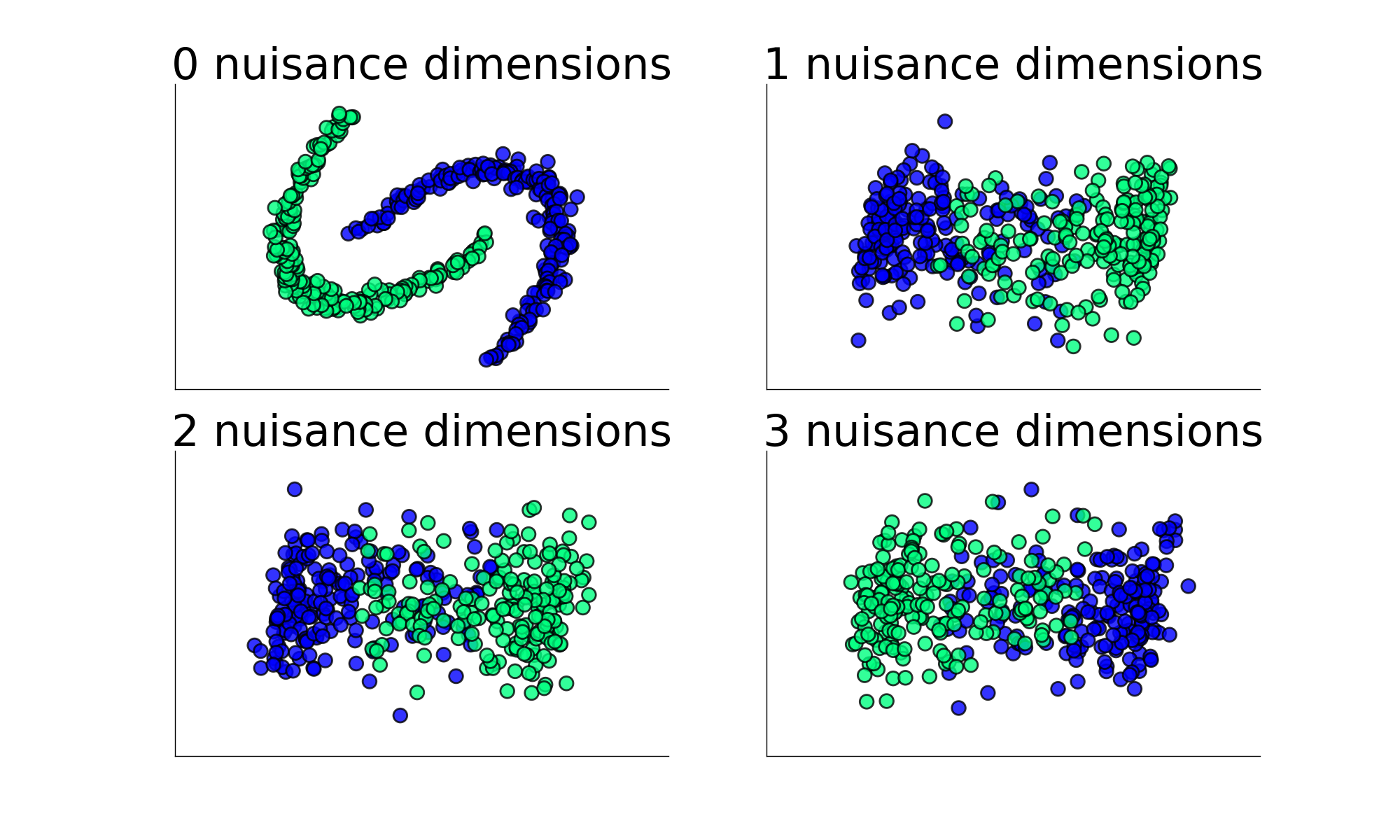}
\caption{Two-moons experiment. Top Left: the original 2-dimensional dataset. Top right: the second largest eigenvalue of the random walk matrix $\myvec{L}_\text{rw}$ decreases as the number $k$ of nuisance dimensions grows. This implies that the graph becomes more connected, and hence less clusterable, as the number of nuisance dimensions grows.
Bottom left: the second largest eigenvector $\myvec{\psi}_2$ (y-axis) of $\myvec{L}_{rw}$ becomes less representative of the true cluster structure as $k$ grows.
Bottom right: projecting the data onto the leading two principal directions (x and y axes) cannot recover the true cluster structure when $k>0$.}
\label{fig:twomoons}
\end{figure*}

\section{Demonstration of the Importance of Unsupervised Feature Selection in High Dimensional Data with Nuisance Features}
\label{sec:motivation}
We first demonstrate the importance of feature selection in unsupervised learning when the data contains nuisance variables, by taking a diffusion perspective. Then, we use a simple two cluster model to analyze how Gaussian nuisance dimensions affect clustering capabilities.
\subsection{A Diffusion Perspective}
Consider the following 2-dimensional dataset, known as ``two-moons'', shown in the top-left panel of Fig.~\ref{fig:twomoons}.
We augment the data with $k$ ``nuisance'' dimensions, where each such dimension is composed of i.i.d $\text{unif}(0,1)$ entries.
As one may expect, when the number of nuisance dimensions is large, the amount of noise (manifested by the nuisance dimensions) dominates the amount of signal (manifested by the two ``true'' dimensions). Consequently, attempts to recover the true structure of the data (say, using manifold learning or clustering) are likely to fail.

From a diffusion perspective, data is considered to be clusterable when the time it takes a random walk starting in one cluster to transition to a point outside the cluster is long. These exit times from different clusters are manifested by the leading eigenvalues of the Laplacian matrix $\myvec{L}_\text{rw}= \myvec{D}^{-1}\myvec{K}$, for which the large eigenvalues (and their corresponding eigenvectors) are the ones that capture different aspects of the data's structure (see, for example,~\cite{nadler2008diffusion}).
Each added nuisance dimension increases the distance between a point and its ``true'' nearest neighbors along one of the ``moons''. In addition, the noise creates spurious similarities between points, regardless of the cluster they belong to. Overall, this shortens the cluster exit times. This phenomenon can be captured by the second largest eigenvalue $\lambda_2$ of $\myvec{L}_\text{rw}$ (the largest eigenvalue $\lambda_1=1$ carries no information as it corresponds to the constant eigenvector $\myvec{\psi}_1$), which decreases as the number of nuisance dimensions grows, as shown in the top right panel of Fig.~\ref{fig:twomoons}. The fact that $\lambda_2$ decreases implies that the diffusion distances~\cite{nadler2008diffusion} in the graph decrease as well, which in turn means that the graph becomes more connected, and hence less clusterable. 
A similar view may be obtained by observing that the second smallest eigenvalue of the un-normalized graph Laplacian $\myvec{L}_\text{un} = \myvec{D} - \myvec{W}$, also known as Fiedler number or algebraic connectivity, grows with the number of nuisance variables. 
The fact that the graph becomes less clusterable as more nuisance dimensions are added is also manifested by the eigenvector $\myvec{\psi}_2$ corresponding to the second largest eigenvalue of $\myvec{L}_\text{rw}$ (or the second smallest eigenvalue of $\myvec{L}_\text{un}$), which becomes less representative of the cluster structure (bottom left panel of Fig.~\ref{fig:twomoons}).

Altogether, this means that in order for the data to be clusterable, the noisy features ought to be removed.
One may argue that principal component analysis can be used to retain the signal features while removing the noise. Unfortunately, as shown in the bottom right panel of Fig.~\ref{fig:twomoons}, projecting the data onto the first two principal directions does not yield the desired result, as the variance along the noise directions is larger than along the signal ones.
In the next sections we will describe our differentiable unsupervised feature selection approach, and demonstrate that it does succeed to recognize the important patterns of the data in this case.
 
\begin{figure*}[tb!]
\centering
\includegraphics[width=0.3\textwidth] {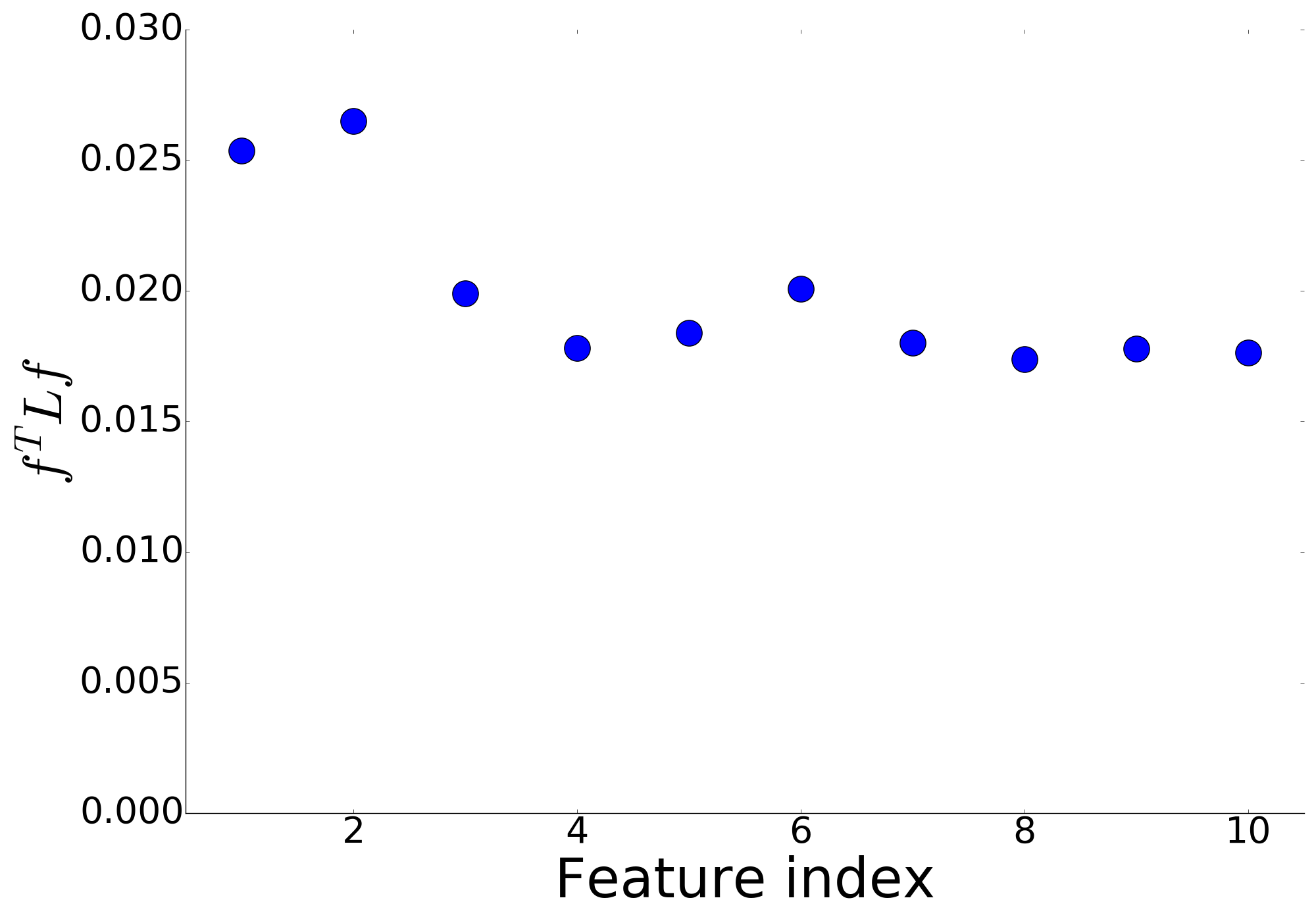}
\includegraphics[width=0.3\textwidth] {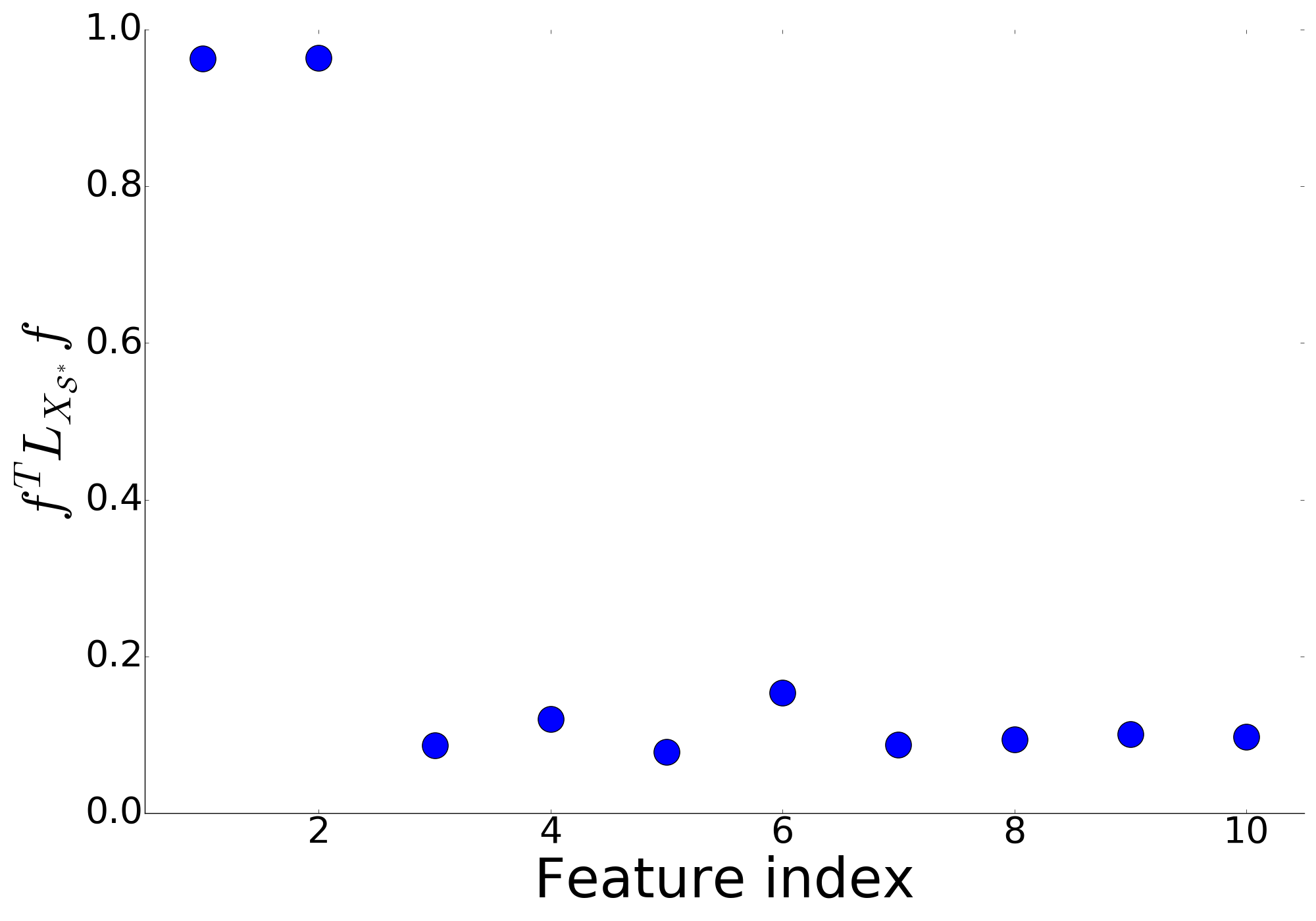}
\includegraphics[width=0.3\textwidth] {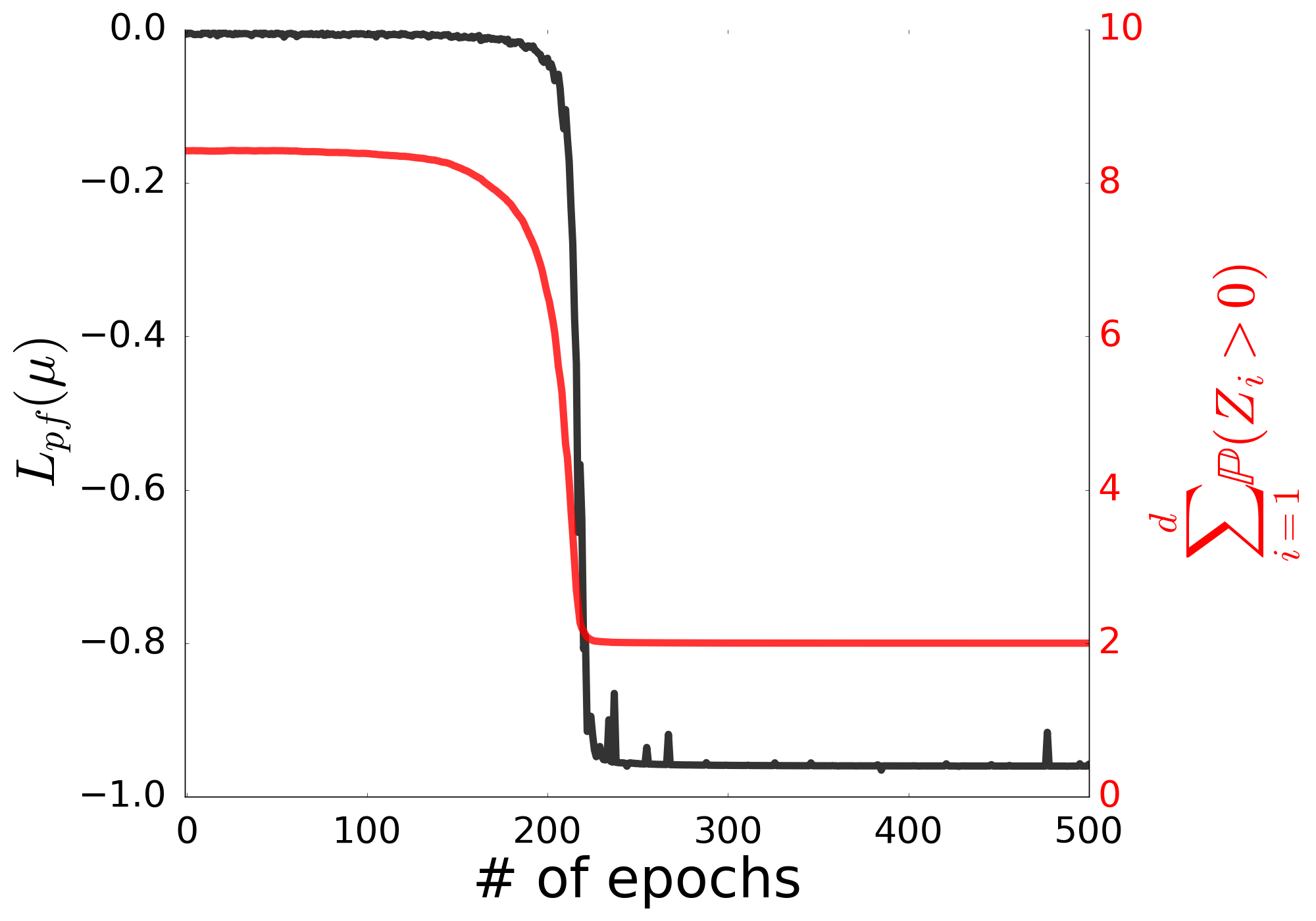}
\caption{Demonstrating the information captured by the Laplacian Score on the noisy-two moons dataset (see Fig. \ref{fig:twomoons}). The first two features are informative, while the rest of the variables are nuisance. Our goal is to train the differentiable stochastic gates for identifying the two informative features.
Left: Laplacian score $\myvec{f}^T \myvec{L} \myvec{f}$ at initialization,  based on all $10$ dimensions in total. The score for the informative features is slightly higher. Middle: Laplacian score, based on the gated Laplacian $\myvec{f}^T \myvec{L}_{{X}_{\cal{S}^*}} \myvec{f}$  at convergence of the gates. The informative features attain a substantial higher score based on the gated Laplacian. Right: the parameter-free loss (black line) and the average number of active gates (red line) as a function of the number of epochs.} 
\label{fig:demo}
\end{figure*}

\subsection{Analysis of Clustering with Nuisance Dimensions}
\label{sec:chi_sq_analysis}
In order to observe the effect of nuisance dimensions, in this section we consider a simple example where all of the noise in the data arises from such dimensions. Specifically, consider a dataset that includes $2n$ datapoints in $\mathbb{R}$, where $n$ of which are at $0 \in \mathbb{R}$ and the remaining ones are at $r >0$, i.e., each cluster is concentrated at a specific point. Next, we add $d$ nuisance dimensions to the data, so that samples lie in $\mathbb{R}^{d+1}$. The value for each datapoint in each nuisance dimension is sampled independently from $N(0, 0.5 ^2)$.

Suppose we construct the graph Laplacian by connecting each point to its nearest neighbors.
We would now investigate the conditions under which the neighbors of each point belong to the correct cluster. 
Consider points $x,y$ belonging to the same cluster. Then $(x-y) = (0, u_1,\ldots, u_d)$ where $u_i\distas{\text{iid}} N(0, 1)$, and therefore $\|x-y\|^2 \sim \chi^2_d$.
Similarly, if $x, y$ belong to different clusters, then  $\|x-y\|^2 \sim r^2 + \chi^2_d$.
Now, to find conditions for $n$ and $d$ under which with high probability the neighbors of each point belong to the same cluster, we can utilize the Chi square measure-concentration bounds~\cite{laurent2000adaptive}.
\begin{lemma}[\cite{laurent2000adaptive} P.1325]
Let $X\sim\chi^2_d$. Then 
\begin{enumerate}
    \item $\mathbb{P}(X-d \ge 2\sqrt{d\gamma}+2\gamma) \le \exp(-\gamma)$.
    \item $\mathbb{P}(d-X \ge 2\sqrt{d\gamma}) \le \exp(-\gamma)$.
\end{enumerate}
\label{lemma:massart}
\end{lemma}

Given sufficiently small $\gamma > 0$ we can divide the segment $[d, d + r^2]$ to two disjoint segments of lengths $2\sqrt{d\gamma}+2\gamma$ and $2\sqrt{d\gamma}$ (and solve for $d$ in order to have the total length $r^2$). 
This yields
\begin{equation}
    \sqrt{d} = \frac{r^2-2\gamma}{4\sqrt{\gamma}}.\label{eq:d}
\end{equation}
The nearest neighbors of each point will be from the same cluster as long as all distances between points from the same cluster will be at most $d + 2\sqrt{d\gamma}+2\gamma$ and all distances between points from different clusters will be at least $d + r^2 - 2\sqrt{d\gamma}$.
According to lemma~\ref{lemma:massart}, this will happen with probability at least $(1 - \exp(-\gamma))^{2n^2 - n}$. 
Denoting this probability as $1 - \epsilon$ and solving for $\gamma$ we obtain
\begin{equation}
    \gamma \le -\log(1-\sqrt[(2n^2 - n)]{1-\epsilon}).\label{eq:k}
\end{equation}
Plugging~\eqref{eq:k} into~\eqref{eq:d} we obtain
\begin{equation}
    d = O\left(\frac{r^4}{-\log(1-\sqrt[(2n^2-n)]{1-\epsilon})}\right).\label{eq:do}
\end{equation}

\begin{figure}[htb!]
\begin{center}
\includegraphics[width=0.45\textwidth] {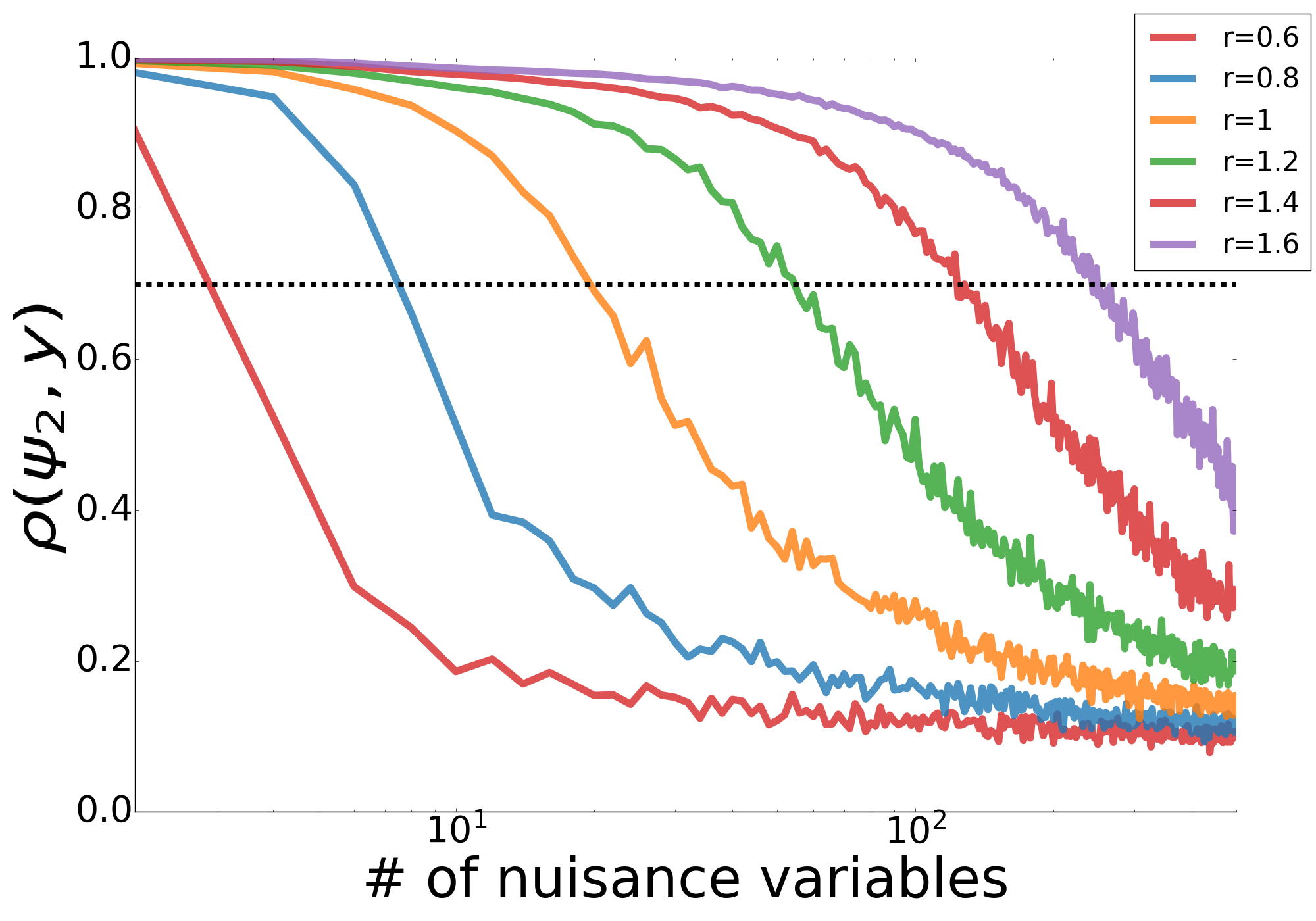}
\includegraphics[width=0.45\textwidth] {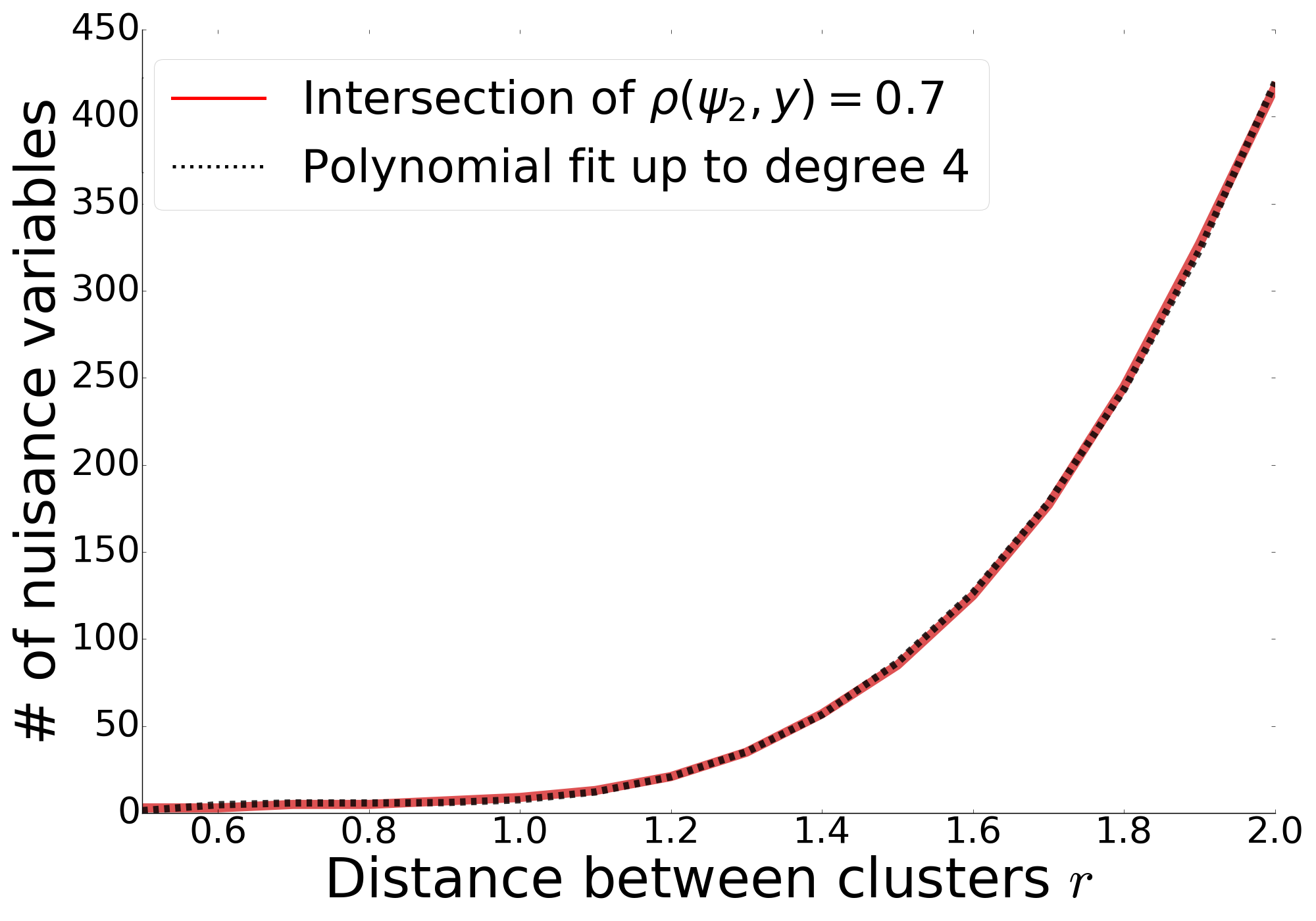}
\end{center}
\caption{Two cluster datasets. We evaluate the influence of Gaussian nuisance variables on the Laplacian. We generate two clusters using $50$ samples each with distance $r$ apart in $1$-D.
We use $d$ Gaussian nuisance variables and evaluate the leading non trivial eigenvector $\psi_2$ of the Laplacian. Left: correlation between the second eigenvector $\psi_1$ and the true cluster assignments $y$ for different values of $r$. As the number of nuisance variables grows, the eigenvector becomes meaningless. As the distance between cluster grows more nuisance variables are required to ``break'' the cluster structure captured by $\psi_2$. Right: by computing the intersection between the damped correlation curves and $0.7$ (shown in the left plot) for different values of $r$ we evaluate the relation between $r$ and number of nuisance variables $d$ required for breaking the cluster structure. This empirical result supports the analysis presented in \ref{sec:chi_sq_analysis} in which we show that $ d = O\left(\frac{r^4}{-\log(1-\sqrt[(2n^2-1)]{1-\epsilon})}\right)$. For convenience we added a polynomial fit up to degree $4$ presented as the black line. } 
\label{fig:chi}
\end{figure}

In particular, for fixed $n$ and $\epsilon$, equation~\eqref{eq:do} implies that the number of nuisance dimensions must be at most on the order of $r^4$ in order for the clusters to not mix with high probability. 
In addition, for a fixed $r$ and $\epsilon$, increasing the number of data points brings the argument inside the log term arbitrarily close to zero, which implies that for large data, the Laplacian is sensitive to the number of nuisance dimensions.
We support these findings via experiments, as shown in Figure~\ref{fig:chi}.

\section{Proposed Method}
\label{sec:proposed}

\begin{figure*}[htb!] 
\centering
\includegraphics[width=0.3\textwidth]{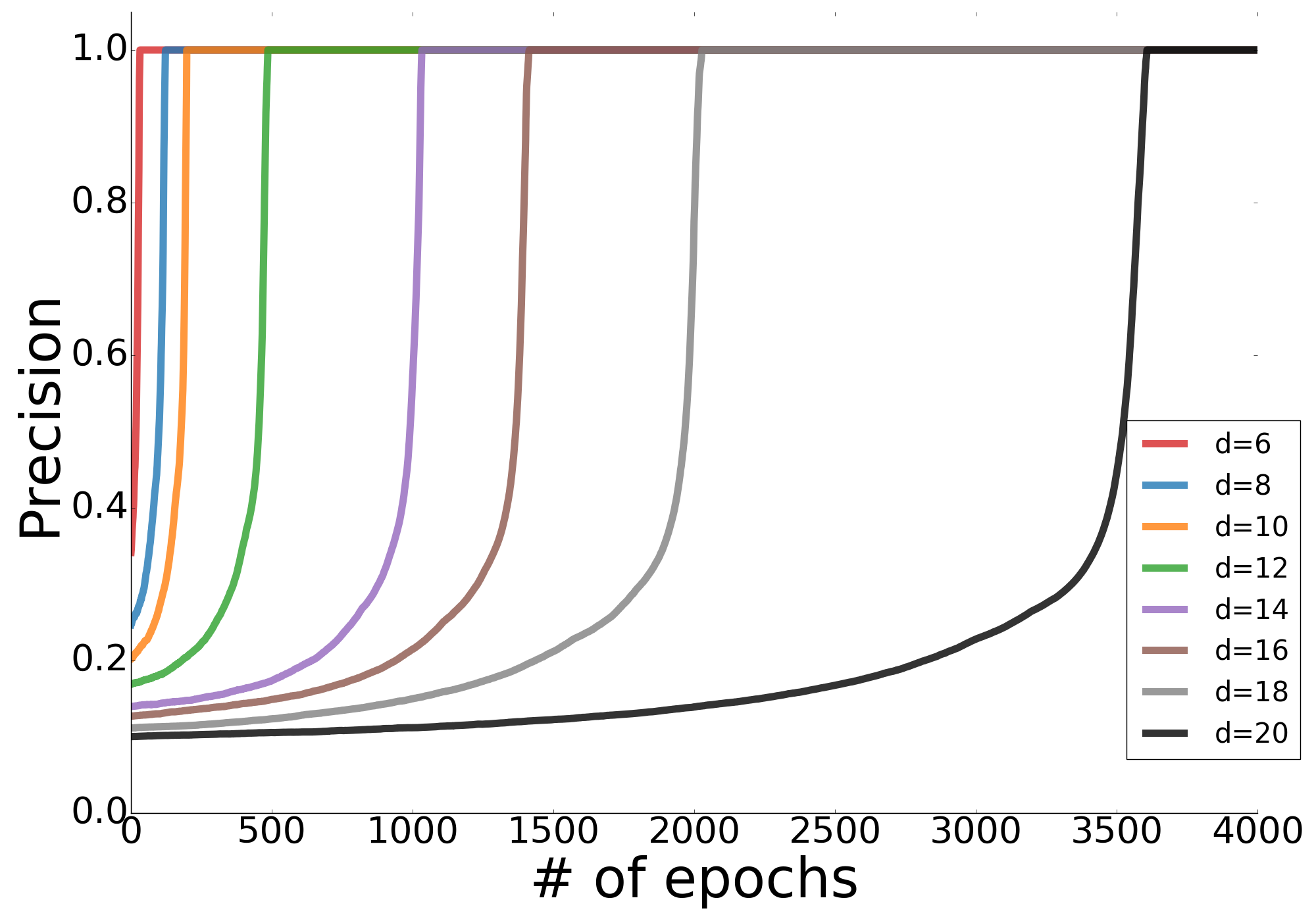}
\includegraphics[width=0.3\textwidth]{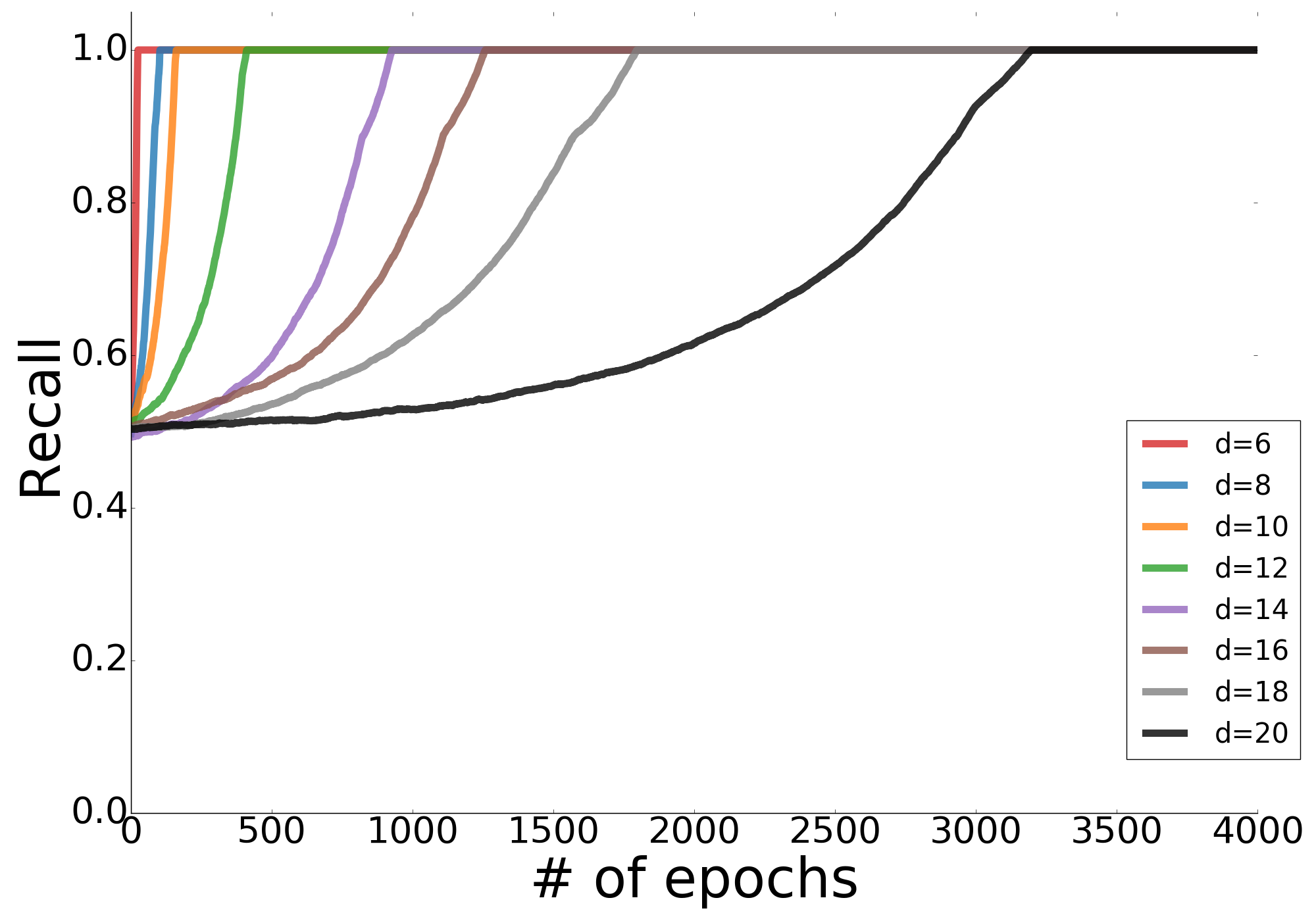}
\includegraphics[width=0.3\textwidth]{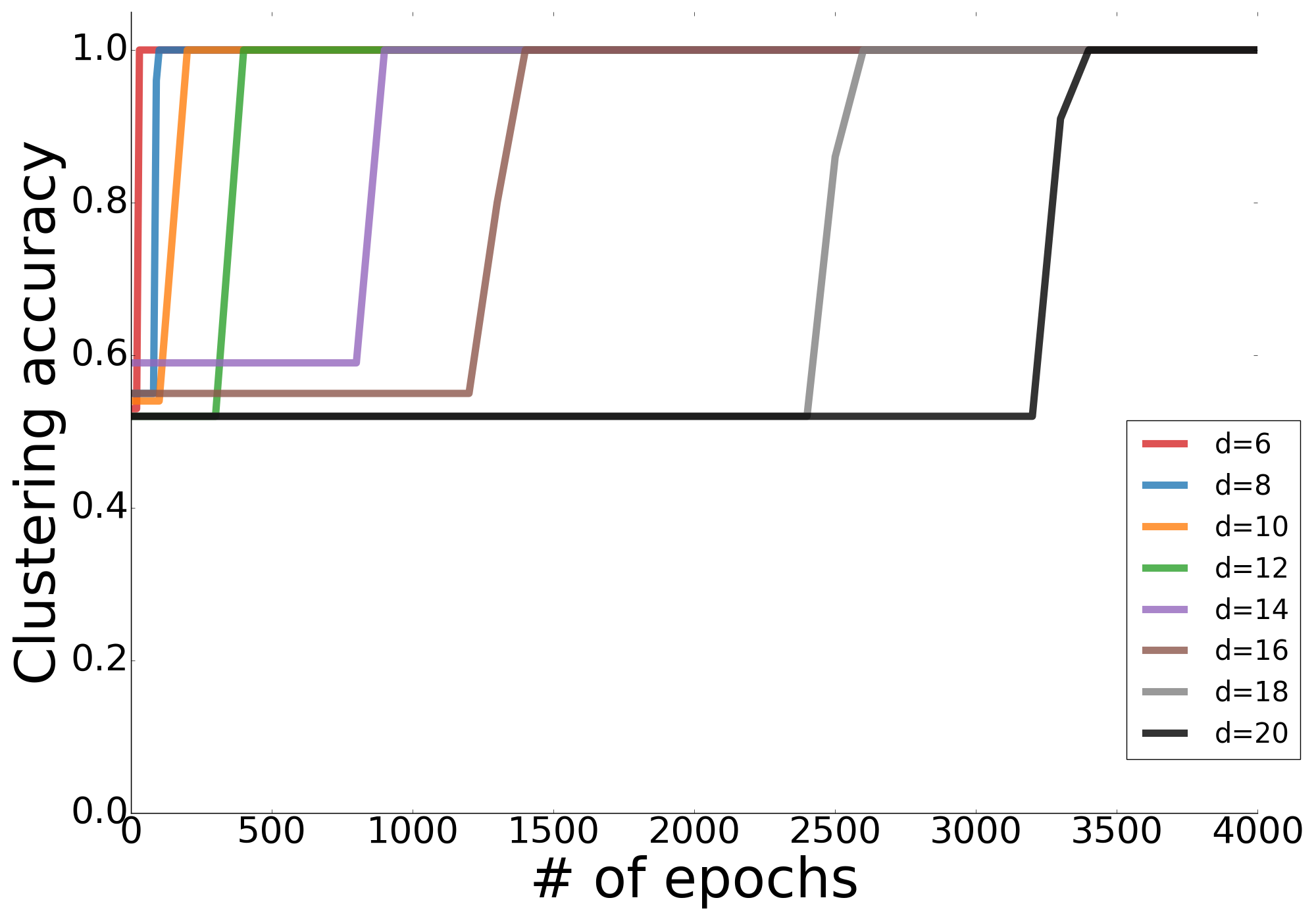}
\caption{Evaluating the precision and recall of feature selection, as well as clustering quality as a function of epoch number. We apply the parameter-free loss variant (see \eqref{eq:param-free}) on the noisy two-moons data with a total of $d$ features. Left: precision of features selection. Here, precision is defined as the ratio between amount of retrieved informative features and all retrieved features, that is $\frac{\sum^2_{i=1}P(Z_i >0)}{\sum^d_{i=1}P(Z_i >0)}$. Middle: recall of features selection. Here, recall is defined as the ratio between amount of retrieved informative features and all informative features, that is $\frac{\sum^2_{i=1}P(Z_i >0)}{2}$. Note that in all of these examples the gates converge to ``deterministic'' values. Namely $P(Z_i >0) \simeq 1$ for informative features $i=1,2$ and $P(Z_i >0) \simeq 0$ for the nuisance features $i=3,...,d$. Right: clustering accuracy obtained with the retrieved features every $10$ epochs. Here, clustering is performed using spectral clustering \cite{ng2002spectral} with a Gaussian kernel.}
\label{fig:lambda_free}
\end{figure*}
\subsection{Rationale}
Recall that the core component of the Laplacian score~\cite{he2006laplacian} is the quadratic term
$\myvec{f}^T \myvec{L} \myvec{f},$
which measures the inner product of the feature $\myvec{f}$ with the eigenvectors of the Laplacian $\myvec{L}$.
For $\myvec{L} = \myvec{L}_{rw} =\myvec{D}^{-1}\myvec{K}$ a large Laplacian score implies that $\myvec{f}$ has a large component in the subspace of eigenvectors corresponding to the largest eigenvalues of $\myvec{L}$. Assuming that the structure of the data varies slowly, these leading eigenvectors (corresponding to large eigenvalues) manifest the main structures in the data, hence a large score implies that a feature contributes to the structure of the data.
However, as we demonstrated in the previous section, 
in the presence of nuisance features, these leading eigenvectors become less representative of the true structure.
In this regime, one could benefit from evaluations of the Laplacian score when the Laplacian is computed based on different subsets $\cal{S}$ of features, i.e., of the form $\myvec{f}^T \myvec{L}_{{X}_{\cal{S}}} \myvec{f},$ where $\myvec{L}_{{X}_{\cal{S}}}$ is the random walk Laplacian computed based on a subset of features $\{ \myvec{f}_{\ell} \}_{\ell \in {\cal{S}}}$. 
Such gated Laplacian score would produce a high score for the informative features $\cal{S}^*$ when the Laplacian is computed only based on these features, that is when $\myvec{L}_{{X}_{\cal{S}}}=\myvec{L}_{{X}_{\cal{S^*}}}$. Searching over all  the different feature subsets is infeasible even for a moderate number of features. Fortunately, we can use continuous stochastic ``gating'' functions to explore the space of feature subsets. 

Specifically, we propose to apply differential stochastic gates to the input features, and compute the Laplacian score after multiplying the input features with the gates. Taking advantage of the fact that informative features are expected to have higher scores than nuisance ones, we penalize open gates. By applying gradient decent to a cost function based on $\myvec{L}_{{X}_{\cal{S}}}$ we obtain the desired dynamic, in which gates corresponding to features that contain high level of noise will gradually close, while gates corresponding to features that are consistent with the true structures in the data will gradually get fully open. This is demonstrated in Fig.~\ref{fig:demo}.

\subsection{Differentiable Unsupervised Feature Selection (DUFS)}

Let $\myvec{X} \in \mathbb{R}^{m\times d}$ be a data minibatch. Let $\myvec{Z}\in [0,1]^d$ be a random variable representing the stochastic gates, parametrized by $\myvec{\mu}\in\mathbb{R}^d$, as defined in Section 2. 
For each mini-batch we draw a vector $z$ of realizations from $\myvec{Z}$ and define a matrix $\tilde{\myvec{Z}}\in [0,1]^{m\times d}$ consisting of $m$ copies of $z$. We denote $\tilde{\myvec{X}} \overset{\triangle}{=} \myvec{X} \odot \tilde{\myvec{Z}}$ as gated input, where $\odot$ is an element-wise multiplication, also known as Hadamard product. Let $\myvec{L}_{\tilde{X}}$ be the random walk graph Laplacian computed on $\tilde{\myvec{X}}$.

 We propose two loss function variants. Both variants contain a feature scoring term $-\frac{1}{m}\mbox{Tr}[\tilde{\myvec{X}} ^T\myvec{L}_{\tilde{X} }\tilde{\myvec{X}}],$ and a feature selection regularization term $\sum_{i=1}^d \mathbb{P}(\myvec{Z}_i \geq 0)$, following~\eqref{eq:reg_term}.
In the first variant~\eqref{eq:param-based} the two terms are balanced using a hyperparameter $\lambda \ge 0$.
\begin{equation}
    L(\mu;\; \lambda):= -\frac{\mbox{Tr} \big[\tilde{\myvec{X}}^T\myvec{L}_{\tilde{X}}\tilde{\myvec{X}} \big]}{m}  + \lambda\sum_{i=1}^d \mathbb{P}(\myvec{Z}_i \ge 0). \label{eq:param-based}
\end{equation}
Controlling $\lambda$ allows for flexibility in the number of selected features. 
To obviate the need to tune $\lambda$, we propose a second loss function, which is parameter-free 
\begin{equation}
    L_\text{param-free}(\mu):= -\frac{\mbox{Tr} \big[ \tilde{\myvec{X}}^T\myvec{L}_{\tilde{X} }\tilde{\myvec{X}} \big]} {m\sum_{i=1}^d \mathbb{P}(\myvec{Z}_i \ge 0) +\delta}\,\, , \label{eq:param-free}
\end{equation}
where $\delta$ is a small constant added to circumvent division by $0$.
The parameter-free variant~\eqref{eq:param-free} seeks to minimize the average score per selected feature, where the average is calculated as the total score (in the numerator) divided by a proxy for the number of selected features (the denominator). Minimizing both proposed objectives \eqref{eq:param-based} and \eqref{eq:param-free} will encourage the gates to remain open for features that yield high Laplacian score, and closed for the remaining features.

Our algorithm involves applying a standard optimization scheme (such as stochastic gradient decent) to objective \eqref{eq:param-based} or \eqref{eq:param-free}. After training, we remove the stochasticity ($\epsilon_i$ in~\eqref{eq:stg}) from the gates and retain features such that $Z_i>0$.

\subsubsection{Raising $\myvec{L}$ to the $t$'th Power}
Replacing the Laplacian $\myvec{L}$ in equations~\eqref{eq:param-based} and~\eqref{eq:param-free} by its $t$-th power $\myvec{L}^t$ with $t > 1$ corresponds to taking $t$ random walk steps \cite{nadler2008diffusion}. This suppresses the smallest eigenvalues of the Laplacian, while preserving its eigenvectors. We used $t=2$, which was observed to improve the performance of our proposed approach (see Appendix for more details).  


\section{Experiments}
To demonstrate the capabilities of DUFS, we begin by presenting an artificial ``two-moons'' experiment. 
We then report results obtained on several standard datasets, and compare them to current baselines. When applying the method to real data we perform feature selection based on Eq.~\eqref{eq:param-based} using several values of $\lambda$. Next, followin the analysis in \cite{wang2015embedded}, we perform clustering using $k$-means based on the leading $50,100,150,200,250,$ or $300$ selected features and average the results over $20$ runs. Leading features are identified by sorting the gates based on $\mathbb{P}({{Z}}_i)$ (see \eqref{eq:reg_term}). The number $k$ of clusters is set as the number of classes and labels are utilized to evaluate clustering accuracy. The best average clustering accuracy is recorded along with the number $|\cal{S}|$ of selected features. 
\label{sec:experiments}

\begin{figure*}[htb!]
\centering
\includegraphics[width=0.4\textwidth] {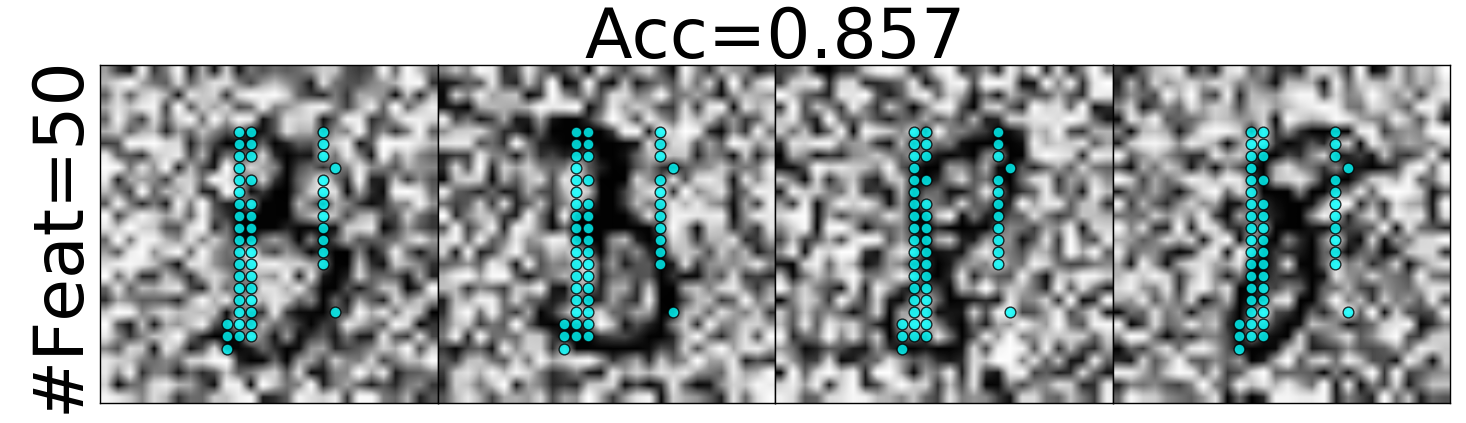}
\includegraphics[width=0.4\textwidth] {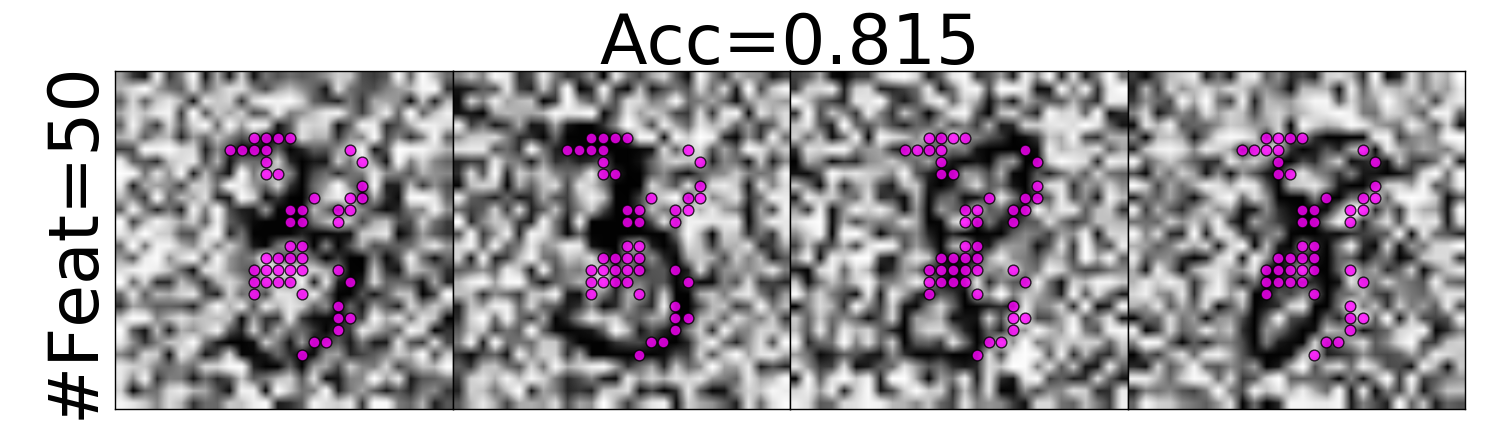}
\includegraphics[width=0.4\textwidth] {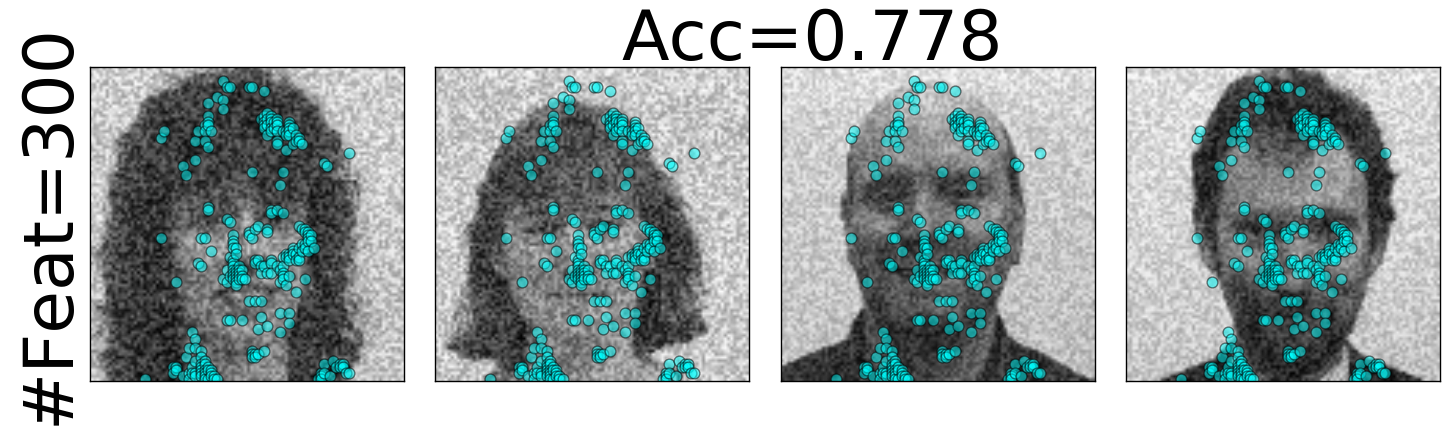}
\includegraphics[width=0.4\textwidth] {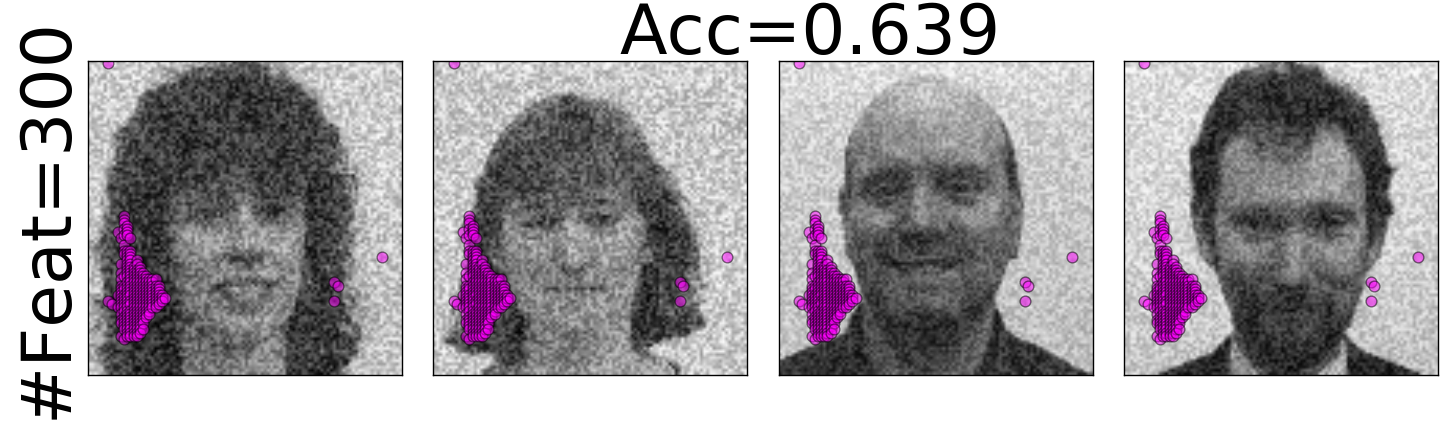}

\caption{Noisy image experiments. Top: examples of noisy MNIST digits highlighted with the leading $50$ features selected by DUFS (left) and LS (right). Bottom: examples from the noisy PIX10 datasets overlaid with the leading $300$ features selected by DUFS (left) and LS (right). This figure is best viewed in color. The gray scale of MNIST images is inverted to improve visibility.} 
\label{fig:mnist}
\end{figure*}

\subsection{Noisy Two Moons}
 \label{sec:twomoon}
In this experiment, we use a two-moons shaped dataset (see Fig.~\ref{fig:twomoons}) concatenated with nuisance features. The first two coordinates $\myvec{f}_1, \myvec{f}_2$ are generated by adding a Gaussian noise with zero mean and variance of $\sigma^2_r=0.1$ onto two nested half circles. Nuisance features $\myvec{f}_i,i=3,...,d$, are drawn from a multivariate Gaussian distribution with zero mean and identity covariance. The total number of samples is $n=100$. Note that the small sample size makes the task of identifying nuisance variables more challenging.

We evaluate the convergence of the parameter-free loss ~\eqref{eq:param-free} using gradient decent. We use different number of features $d$ and plot the precision and recall of feature selection throughout training (see Fig. \ref{fig:lambda_free}). In all of the presented examples, perfect precision and recall are achieved at convergence.


\begin{figure*}[htb!] 
\centering
\includegraphics[width=0.32\textwidth]{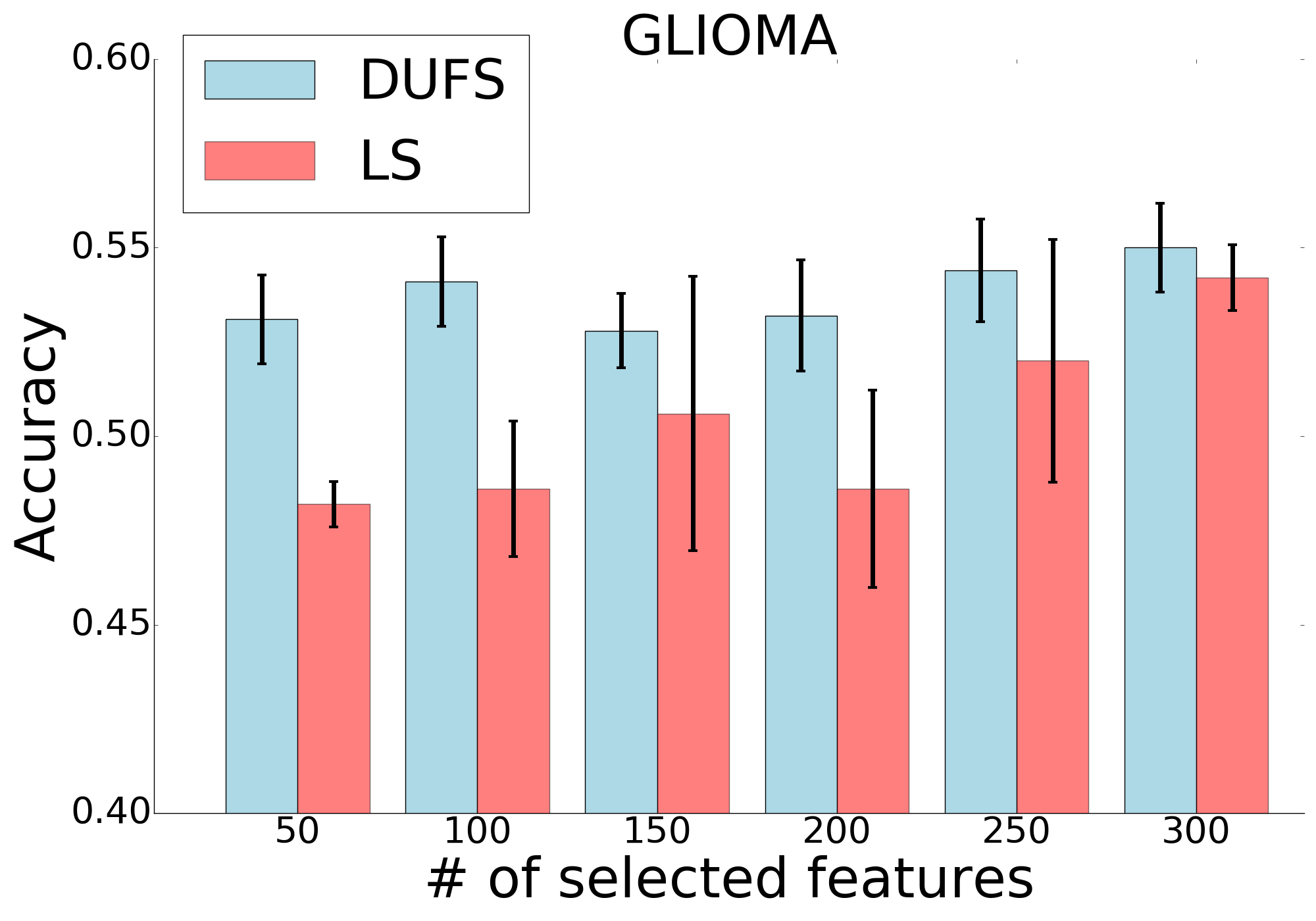}
\includegraphics[width=0.32\textwidth]{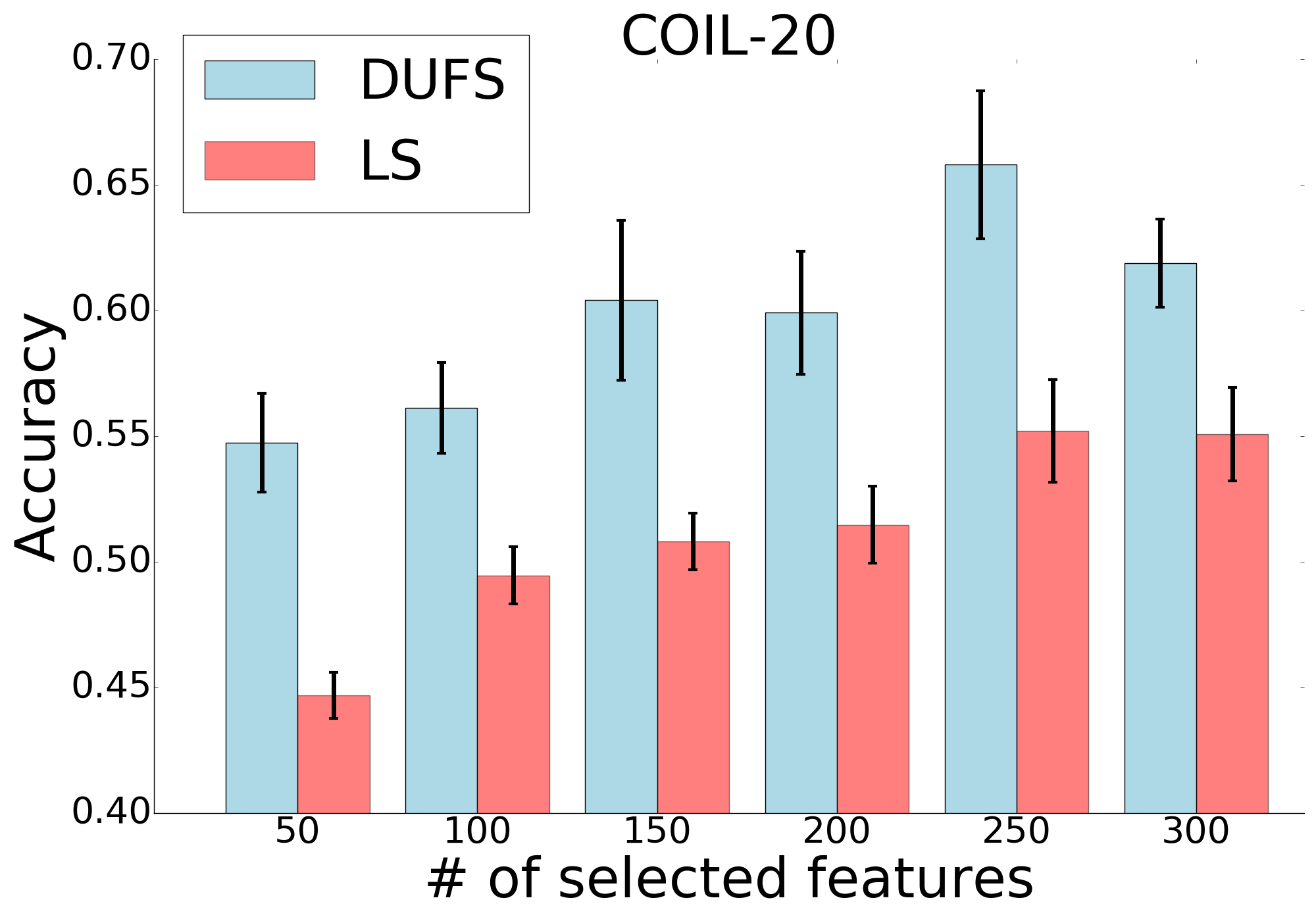}
\includegraphics[width=0.32\textwidth]{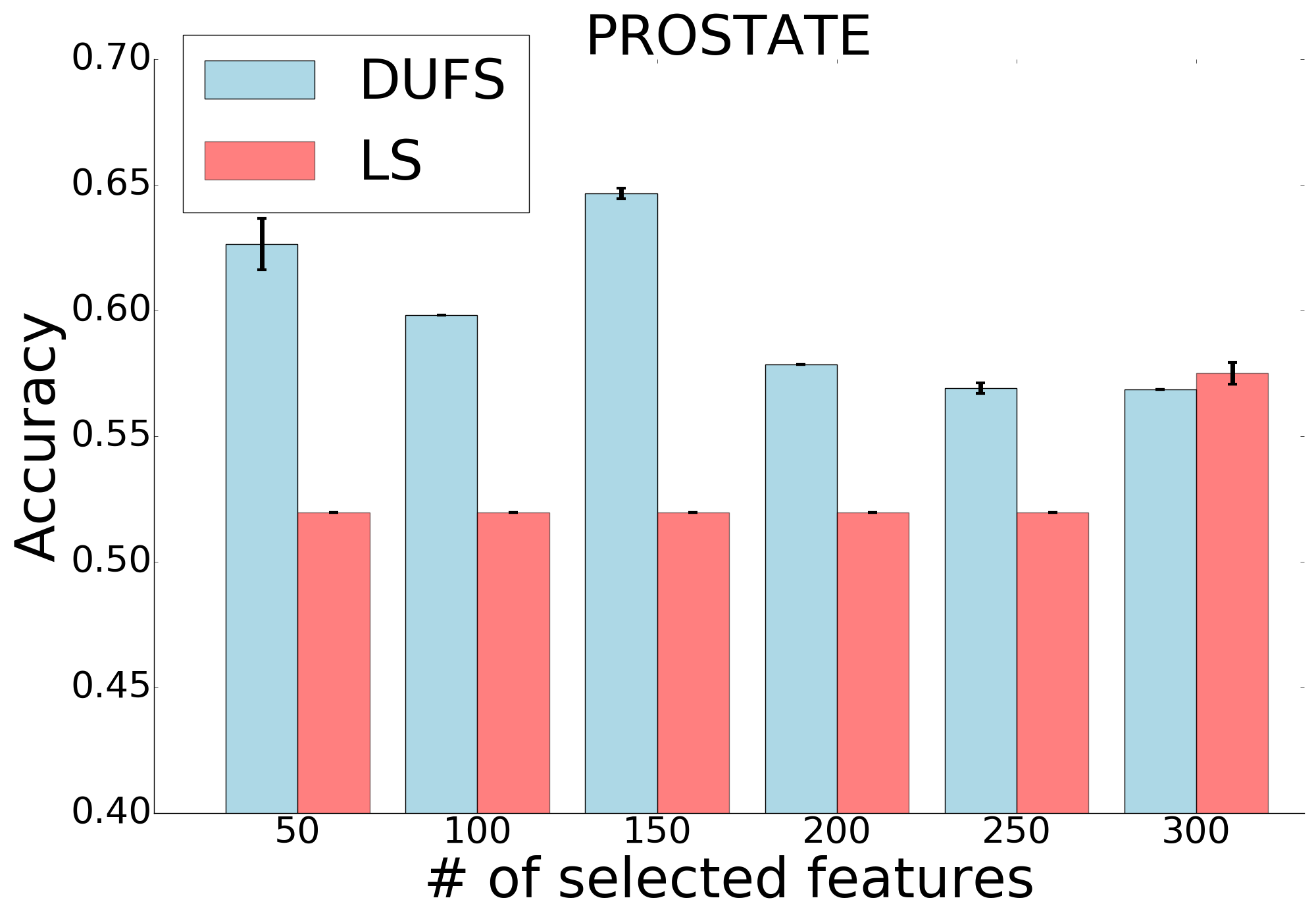}
\caption{Clustering accuracy on three real world datasets. Clustering was performed by applying $k$-means to features selected by DUFS and LS. The averages and standard deviations based on $20$ runs are shown. }
\label{fig:real_data}
\end{figure*}

\subsection{Noisy Image Data}
\label{sec:mnist}

In the following experiment we evaluate our method on two noisy image datasets. The first is a noisy variant of MNIST~\cite{mnist}, in which each background pixel is replaced by a random value drawn uniformly from $[0,1]$ (see also~\cite{qi2019activity}). 
Here we focus only on the digits `3` and `8`. The second dataset is a noisy variant of PIXRAW10P (abbreviated PIX10), created by adding noise drawn uniformly from $[0,0.3]$ to all pixels. In both datasets  the images were scaled into $[0,1]$ prior to the addition of noise.
We applied to both datasets the DUFS and LS approaches.
In the top panels of Fig.~\ref{fig:mnist} we present the leading $50$ features retained on noisy MNIST along with the average clustering accuracy over $20$ runs of $k$-means. 
In this case, DUFS' open gates concentrate at the left side of the handwriting area, which is the side that distinguishes `3` from `8`. This allows DUFS to achieve a higher clustering accuracy comparing to LS . We remark that the training was purely unsupervised and all label information was absent. 
The bottom panels of Fig.~\ref{fig:mnist} show the leading $300$ features retained on noisy PIX10 along with the average clustering accuracy. Here, DUFS selects features which are more informative for clustering the face images. We refer the reader to Appendix~\ref{sec:image} for additional information on the noisy image datasets along with extended results on noisy images.

\subsection{Clustering of Real World Data}
Here, we evaluate the capabilities of the proposed approach (DUFS) in clustering of real datasets. We use several benchmark datasets\footnote{http://featureselection.asu.edu/datasets.php}\footnote{For RCV1 we use a binary subset (see exact details in the Appendix) of the full data: https://scikit-learn.org/0.18/datasets/rcv1.html}. The properties of all datasets are summarized in Table \ref{tab:results}. We compare DUFS to Laplacian Score \cite{belkin2002laplacian} (LS), Multi-Cluster Feature Selection \cite{cai2010unsupervised} (MCFS), Local Learning based Clustering (LLCFS) \cite{zeng2010feature}, Nonnegative Discriminative Feature Selection (NDFS) \cite{li2012unsupervised}, Multi-Subspace 
Randomization and Collaboration (SRCFS) \cite{huang2019unsupervised} and Concrete Auto-encoders (CAE) \cite{CAE}. In table \ref{tab:results} we present the accuracy of clustering based on feature selected by DUFS, the 6 baselines and based on all features (All). 
As can be seen, DUFS outperforms all baselines on $6$ datasets, and is ranked at second place on the remaining $3$. Overall the median and mean ranking of DUFS are $1$ and $1.3$.


\begin{table*}[b]
  \centering
  \begin{adjustbox}{width=1 \columnwidth,center}
  \small{
    \begin{tabular}{lllllllll|r }
        \toprule
    Datasets & LS    & MCFS  & NDFS  & LLCFS & SRCFS  & CAE  &  DUFS  & All  &   Dim/Samples/Classes\\
      \midrule
     GISETTE & 75.8 (50) & 56.5 (50) & 69.3 (250) &  72.5 (50) & 68.5 (50) & 77.3 (250) & {\bf{99.5 (50)}}     & 74.4  & 4955 / 6000 / 2\\
    PIX10 & 76.6 (150) & 75.9 (200) & 76.7 (200)&  69.1 (300) & 76.0 (300) & {\bf{94.1 (250)} }& 88.4 (50)     & 74.3  & 10000 / 100 / 10\\
    COIL20 & 55.2 (250) & 59.7 (250) & 60.1 (300) & 48.1 (300) & 59.9 (300) & 65.6 (200) & {\bf{ 65.8 (250)}}     & 53.6 & 1024 / 1444 / 20\\
    Yale  & 42.7 (300) & 41.7 (300)  & 42.5 (300) & 42.6 (300) & 46.3 (250) & 45.4 (250) & {\bf{ 47.9 (200)} }     & 38.3 & 1024 / 165 / 15\\
    TOX-171 & 47.5 (200) & 42.5 (100) & 46.1 (100) & 46.7 (250) & 45.8 (150) &  44.4 (150) &   \bf{49.1 (50)}    & 41.5& 2000 / 62 / 2\\
    ALLAML & 73.2 (150) & 68.4 (100) & 69.4 (100) & {\bf{ 77.8 (50)}} & 67.7 (250) & 72.2 (200) &  74.5 (100)     & 67.3& 7192 / 72 / 2\\
    PROSTATE & 57.5 (300) & 57.3 (300) & 58.3 (100) & 57.8 (50) &60.6 (50) & 56.9 (250) & {\bf{ 64.7 (150)}}     & 58.1 & 5966 / 102 / 2\\
        RCV1 & 54.9 (300) & 50.1 (150) & 55.1 (150) & 55.0 (300) & 53.7 (300) &  54.9 (300) &  {\bf{ 60.2 (300)}}   & 50.0 & 47236 / 21232 /  2\\
        ISOLET & 56.4 (300) & 62.1 (300) & 60.4 (100) & 52.3 (100) & 60.1 (250) & \bf{63.8 (250)} & 63.6 (150) & 60.8& 617 / 1560 / 26\\
        \midrule
        Median rank & 4  & 6 & 4 & 4 & 5 & 3 & \bf{1} & \\
        Mean rank & 4.1  & 6 & 3.9 & 4.6 & 4.7 & 3.4 &  \bf{1.3}&  \\
        \bottomrule
    \end{tabular}%
    }
\end{adjustbox}

  \caption{Left table- Average clustering accuracy on several benchmark datasets. Clustering is performed by applying $k$-means to the features selected by the different methods. The number of selected features is shown in parenthesis.  Right table- Properties of the real world data used for empirical evaluation.} 
   \label{tab:results}

\end{table*}%

In the next experiment we evaluate the effectiveness of the proposed method for different numbers of selected features on $3$ datasets. We compare DUFS versus LS by performing $k$-means clustering using the features selected by each method. 
In Fig. \ref{fig:real_data} we present the clustering accuracies (averaged over $20$ runs) based on the leading $\{50,100,...,300\}$ features. We see that DUFS consistently selects features which provide higher clustering capabilities compared to LS.

\section{Conclusions}
\label{sec:conclusion}

In this paper, we propose DUFS, a novel unsupervised feature selection method by introducing learnable Bernoulli gates into a Laplacian score. DUFS has an advantage over the standard Laplacian score as it re-evaluates the Laplacian score based on the subset of selected features.
We demonstrate that our proposed approach captures structures in the data that are not detected by mining the data with standard Laplacian, in the presence of nuisance features. Finally, we experimentally demonstrate that our method outperforms current unsupervised feature selection baselines on several real-world datasets.

\section*{Acknowledgements}
The authors thank Stefan Steinerberger, Boaz Nadler and Ronen Basri for helpful discussions. This work was supported by the National Institutes of Health [R01GM131642, UM1DA051410, P50CA121974 and R61DA047037].

\bibliographystyle{ieeetr}
\bibliography{biblio}

\clearpage

\beginsupplement
\section*{Appendix}

\section{Tuning the Kernel's Bandwidth}

It is important to properly tune the kernel scale/bandwidth $\sigma_b$, which determines the scale of connectivity of the kernel $\myvec{K}$. 
Several studies have proposed schemes for tuning $\sigma_b$, see for example \cite{Amit2,Keller,zelnik,epsilon}. 
Here, we focus on two schemes, a global bandwidth and a local bandwidth. The local bandwidth proposed in \cite{zelnik}, involves setting a local-scale $\sigma_i$ for each data point $\myvec{x}_i,i=1,...,n$. The scale is chosen using the $L_1$ distance from the $k$-th nearest neighbor of the point $\myvec{x}_i$. Explicitly, the calculation for each point is 
\begin{equation}\label{eq:epsilon}
\sigma_i=C \cdot ||\myvec{x}_i-\myvec{x}_k||^2,i=1,...,N,
\end{equation} where $\myvec{x}_k$ is the $k$-th nearest (Euclidean) neighbor of the point $\myvec{x}_i$, and $C$ is a predefined constant in the range $[1,5]$. We compute $\hat{\sigma}_b$ as the max over $\sigma_i$, then, the value of the kernel for points $\myvec{x}_i \text{ and } \myvec{x}_j$ is
\begin{equation}\label{GKernelZelnik}
 K_{i,j}=\exp\left( {-\frac{||\myvec{x}_i-\myvec{x}_j||^2}{
	\hat{\sigma}_b}  }\right),i,j\in\{1\ldots n\}.
\end{equation} This scale guarantees that all of the points are connected to at least $k$ neighbors. 
\section{Feature selection on Image Datasets}
\label{sec:image}
\begin{figure}[h]
\begin{center}
\subfigure[DUFS]{\includegraphics[width=0.85\textwidth] {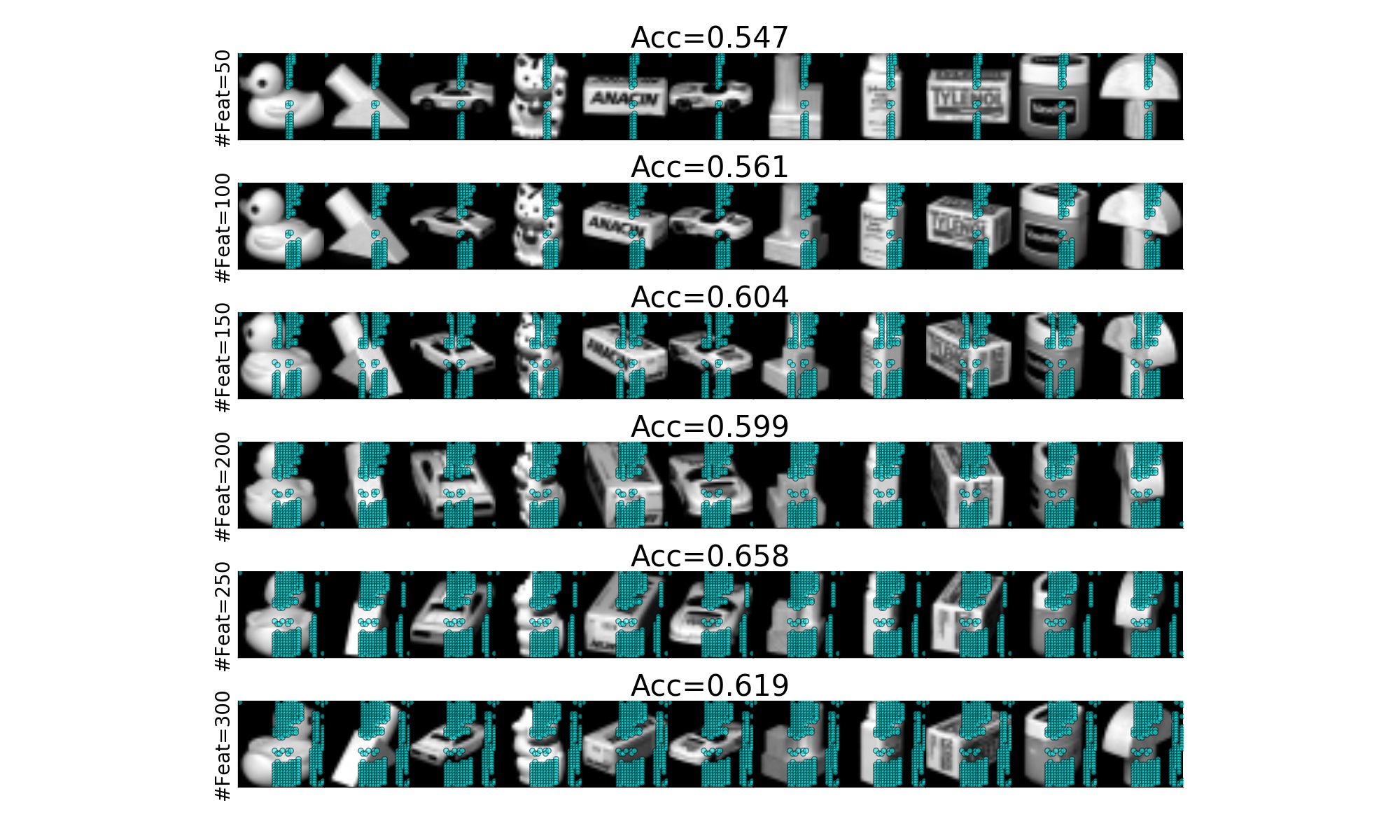}}
\subfigure[LS]{\includegraphics[width=0.85\textwidth] {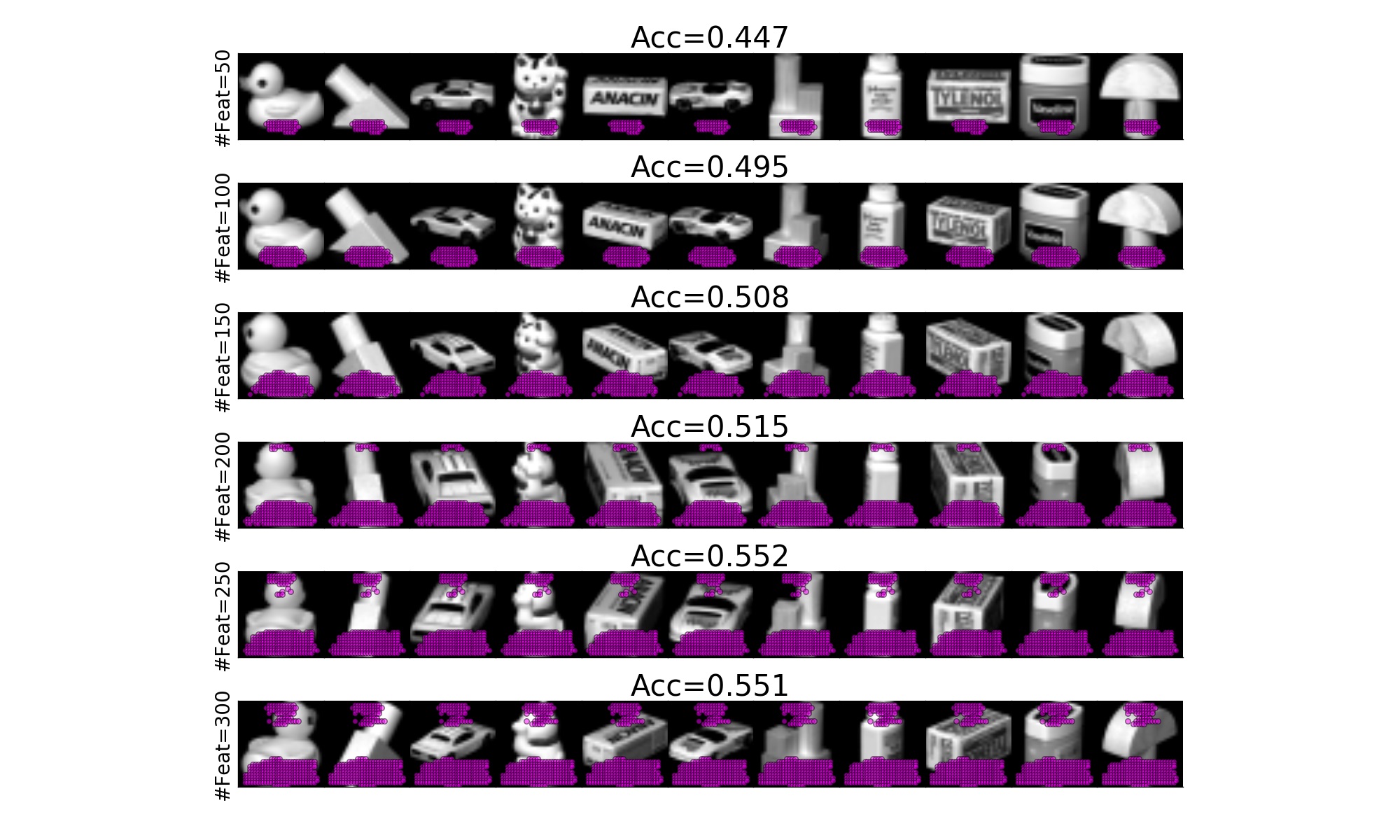}}
\end{center}
\caption{Features selected by DUFS and LS in the COIL20 dataset. Top: selected features (cyan dots) and clustering accuracy based on DUFS. Note that as COIL20 contains different angles of each object, the selected feature lie approximately on the symmetry axis. Bottom: selected features (magenta dots) and clustering accuracy based on LS.} 
\label{fig:coil20}
\end{figure}

\begin{figure}[h]
\begin{center}
\subfigure[DUFS]{\includegraphics[width=0.85\textwidth] {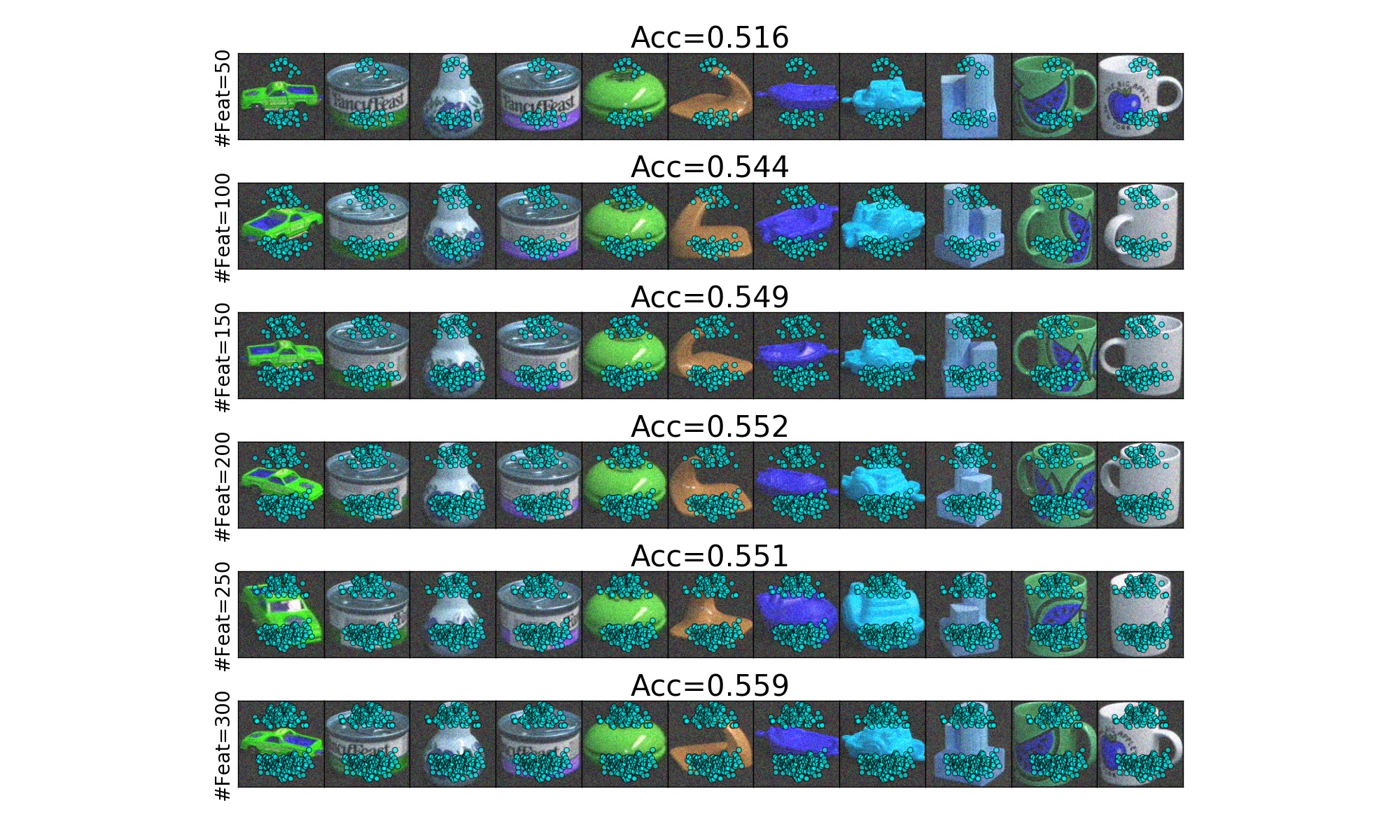}}
\subfigure[LS]{\includegraphics[width=0.85\textwidth] {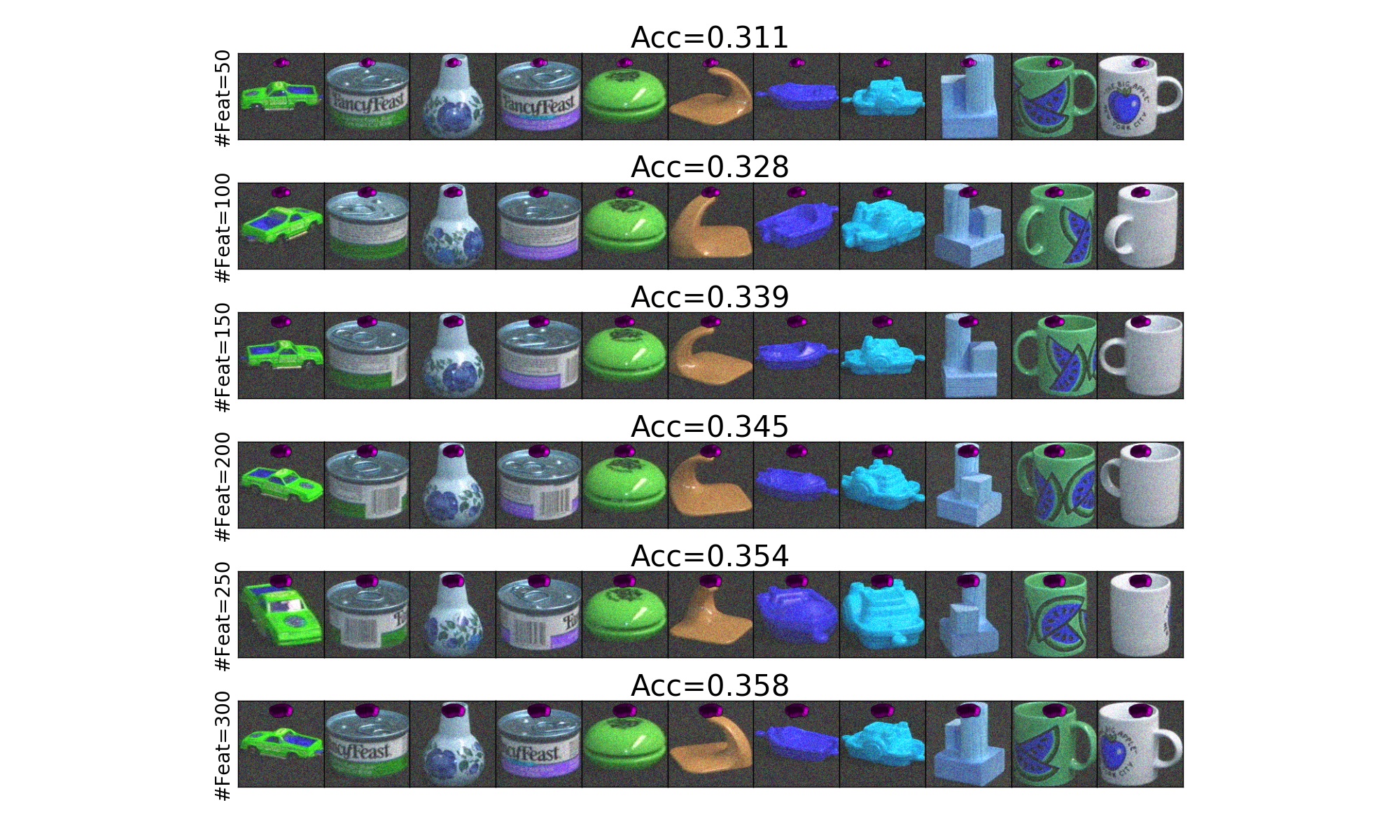}}
\end{center}
\caption{Same as for \ref{fig:coil20} but for the COIL100 dataset. Note that as COIL100 contains different angles of each object, the selected feature lie approximately on the symmetry axis. In this example, the LS also selects features on the symmetry axis, however the LS based selected features are condensed at a small region near the top part of the image. These features are informative for clustering wide vs. long objects but less informative for clustering all $100$ objects.} 
\label{fig:coil100}
\end{figure}

Here, we provide a deeper look into the features identified by the proposed method when applied to image data. We start with COIL20 which is a data that contains $20$ objects captured at different viewing angles. In Fig. \ref{fig:coil20} we present the leading $\{50,100,...,300\}$ features selected by DUFS and LS along with the average clustering accuaracies based on the selected features. In this example DUFS selects features which lie on the symmetry axis of COIL20, these features are more informative for clustering COIL20 since the values of rotated objects vary slowly on this axis. Next, we present a similar comparison on COIL100. COIL100 contains $7200$ samples of $100$ objects captured at different angles. Each image is of dimension $[128, 128, 3]$. In Fig. \ref{fig:coil100} we present the leading $\{50,100,...,300\}$ features selected by DUFS and LS along with the average clustering accuracies based on the selected features. Here, feature selection is performed based on a black and white version of the RGB image and clustering is performed based on the corresponding subset of pixels from the RGB tensor. 

\begin{figure}[h]
\begin{center}
\subfigure[DUFS]{\includegraphics[width=0.95\textwidth] {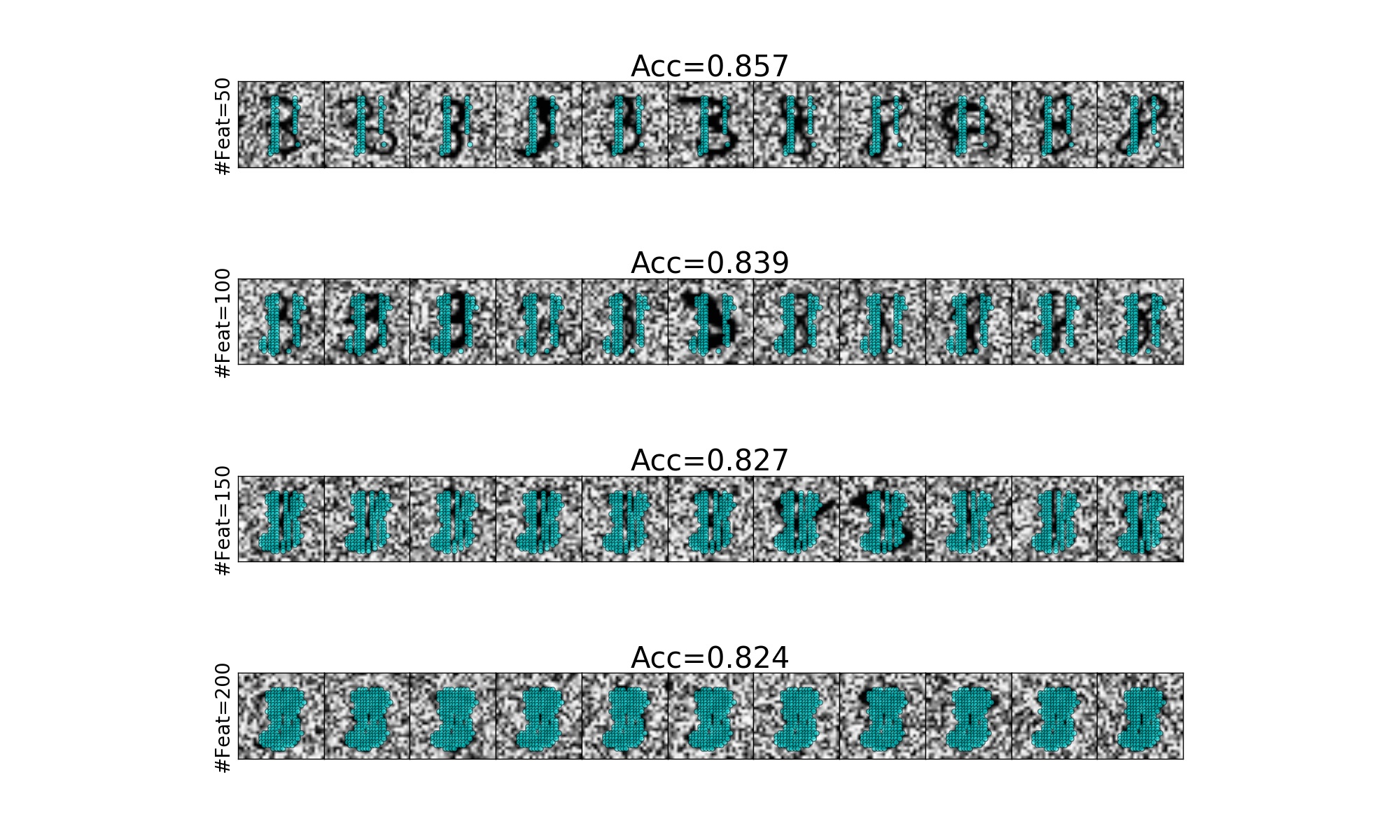}}
\subfigure[LS]{
\includegraphics[width=0.95\textwidth] {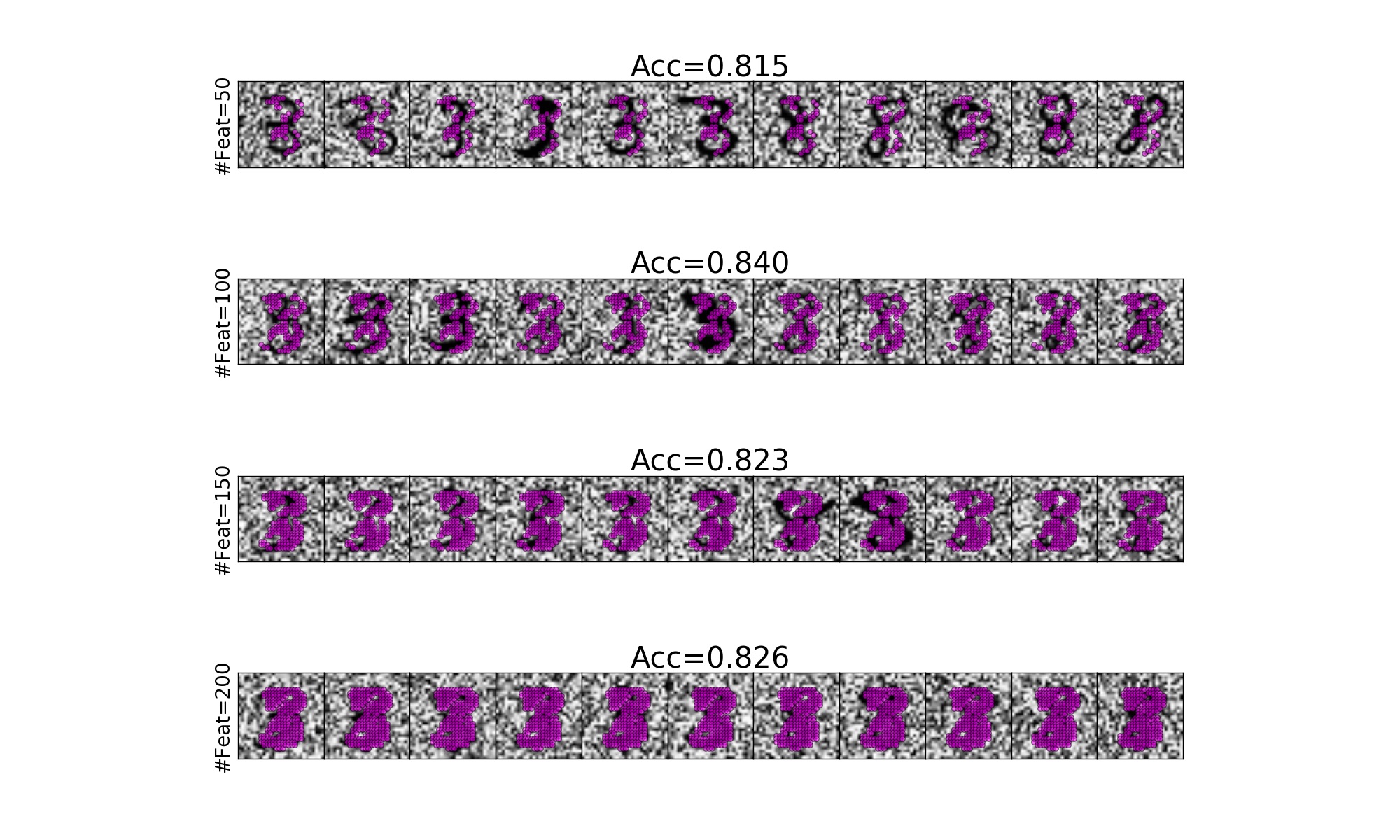}}
\end{center}
\caption{Selected features on MNIST dataset. Top: selected features and clustering accuracy based on DUFS. Bottom: selected features and clustering accuracy based on LS. In this example, DUFS outperforms the LS when it is regularized to select a small number of features. However, when the regularization is set to select $>100$ features in DUFS, the features with top scores in both methods are similar.} 
\label{fig:mnistfull}
\end{figure}

Finally, in Fig. \ref{fig:mnistfull} we present the results of application of DUFS to the noisy MNIST dataset. This is an extension of the results presented in the paper. Specifically, we demonstrate the clustering accuracies based on the leading $\{50,100,...,300\}$ features selected by DUFS and LS. In this experiment, we focused on a random subset of $1000$ samples of the digits $3$ and $8$.

\section{Raising $\myvec{L}$ to the $t$'th Power}
\begin{figure}[h] 
\centering

\includegraphics[width=0.8\textwidth]{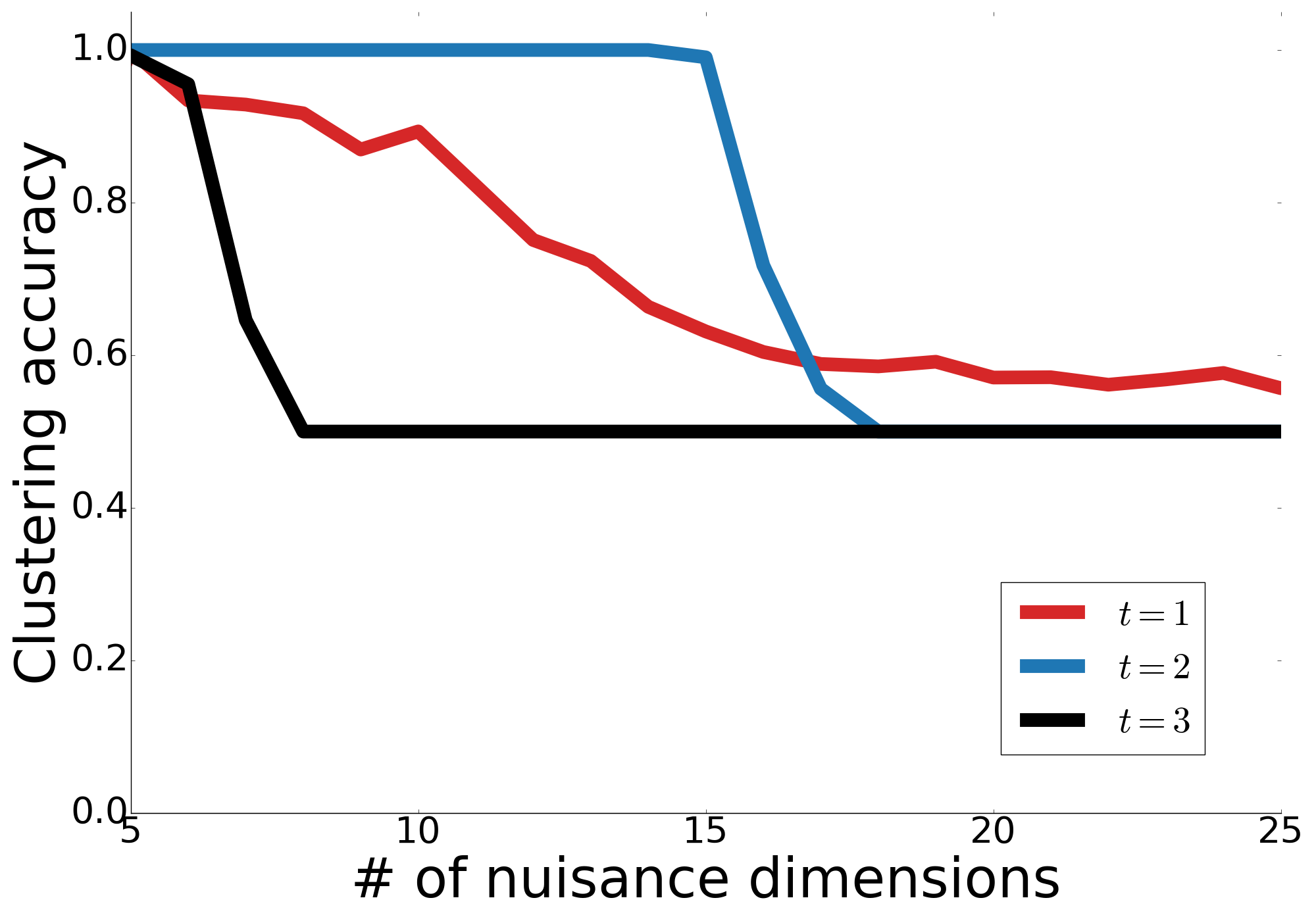}
\caption{Taking powers of the graph Laplacian. Clustering accuracy vs. number of nuisance dimensions in the two-moons datasts. We apply the parameter free variant of DUFS using a Laplacian $\myvec{L}$ raised to the power of $t$. Clustering is performed using $k$-means applied to the selected features and averaged over $100$ runs of DUFS.}
\label{fig:power}
\end{figure}

To suppress the smallest eigenvalues of the Laplacian, we have suggested to replace the Laplacian $\myvec{L}$ in equations~\eqref{eq:param-based} and~\eqref{eq:param-free} by its $t$-th power $\myvec{L}^t$ with $t > 1$. As shown in \cite{nadler2008diffusion} this corresponds for taking $t$ random walk steps on the graph of the data. In this subsection we empirically demonstrate the effect of $t$ using the two-moons dataset (described in the Experimental section of the main text). We construct the two-moons dataset with different number of nuisance variables ($d$) and apply DUFS (with the parameter free loss) computed based on $\myvec{L}$ raised to the power of $t=1,2$ and $3$. In Fig. \ref{fig:power} we present the clustering accuracy (averaged over $100$ runs) based on $k$-means, which is applied to the selected features. As evident in this plot the Laplacian based on $t=2$ yields better performance for a wider range of nuisance variables in this experiment. Following this result, we keep $t=2$ in all of our examples.

\section{Experimental Details}

In this subsection we describe all required details for conducting the experiments provided in the main text. We use SGD for all the experiments which are conducted using Intel(R) Xeon(R) CPU E5-2620 v3 @2.4Ghz x2 (12 cores total). For LS, MCFS, and NDFS we use a python implementation from \footnote{https://github.com/jundongl/scikit-feature}. For LLCFS and SRCFS we use a Matlab implementation from \footnote{https://github.com/huangdonghere/SRCFS} and \footnote{https://www.mathworks.com/matlabcentral/fileexchange/56937-feature-selection-library}. For CAE we use a python implementation available at \footnote{https://github.com/mfbalin/Concrete-Autoencoders}. For DUFS and LS, we use $k=2$ (number of nearest neighbors) which worked well on all of the datastes, except for ISOLET and GISSETE datasets in which we used $k=5$. The factor $C$ (see Eq. \ref{eq:epsilon}) in all experiments is $5$ except for COIL20 and PIX10 in which $C=2$. All datasets are publicly available at \footnote{http://featureselection.asu.edu/datasets.php}, except RCV1 which is available at \footnote{https://scikit-learn.org/0.18/datasets/rcv1.html}. RCV1 is a multi-class multi-label datasets, in our analysis we use a binary subset of RCV1. To create this subset, we focus on the first two classes and remove all samples that have multiple labels, then we balance the classes by down sampling the larger class. For NDFS, MCFS and SRCFS we use $k=5$ for the affinity matrix $\myvec{W}$, note that NDFS and LLCFS use the number of clusters for selecting features. The tuning process for hyper-parameters of all method follows the grid search described described in \cite{wang2015embedded}.

In all examples except RCV1, COIL100 and COIL20 we use a full batch size for computing the kernel, for COIL100 and COIL20 the batch size is $1000$. For all two-moons examples presented in Fig. \ref{fig:twomoons} we use the parameter free loss term with a learning rate (LR) of $1$ and $5000$ epochs. For PROSTATE data, we use a learning rate of $1$, $12000$ epochs and $\lambda$ is evaluated in the range $[0.01,1]$. For GLIOMA data we use a learning rate of $0.3$, $12000$ epochs and $\lambda$ is evaluated in the range $[3,30]$. For ALLAML data we use a learning rate of $0.3$, $20000$ epochs and $\lambda$ is evaluated in the range $[1,5]$. For COIL20 data we use a learning rate of $0.3$, $26000$ epochs and $\lambda$ is evaluated in the range $[0.01,2]$. For COIL100 data we use a learning rate of $1$, $6000$ epochs and $\lambda$ is evaluated in the range $[0.01,2]$. For PIX10 data we use a learning rate of $0.3$, $20000$ epochs and $\lambda$ is evaluated in the range $[0.05,1]$. For ISOLET data we use a learning rate of $0.3$, $20000$ epochs and $\lambda$ is evaluated in the range $[0.1,10]$.

\end{document}